\RequirePackage{fix-cm}
\documentclass[twocolumn]{svjour3}          
\smartqed  
\usepackage{graphicx}
\usepackage{epstopdf}
\usepackage{amsmath}
\usepackage{natbib}
\usepackage{stfloats}
\newcommand{\tabincell}[2]{\begin{tabular}{@{}#1@{}}#2\end{tabular}}
\begin{document}

\title{Video Primal Sketch
}
\subtitle{A Unified Middle-Level Representation for Video}


\author{Zhi Han$^{1,2,3}$         \and
        Zongben Xu$^{1}$      \and
        Song-Chun Zhu$^{2}$
}


\institute{1, Institute for Information and System Sciences, Xi'an Jiaotong University, China.\\
           2, Dept. of Stat and CS, University of California, LA, USA.\\
           3, State Key Laboratory of Robotics, Shenyang Institute of Automation, Chinese Academy of Sciences, China.\\
             \email{han.zhi.hunt@gmail.com; zbxu@mail.xjtu.edu.cn; sczhu@stat.ucla.edu}
}

\date{Received: date / Accepted: date}

\maketitle

\begin{abstract}
This paper presents a middle-level video representation named Video Primal Sketch (VPS), which integrates two regimes of models: i) sparse coding model using static or moving primitives to explicitly represent moving corners, lines, feature points, etc., ii) FRAME /MRF model reproducing feature statistics extracted from input video to implicitly represent textured motion, such as water and fire. The feature statistics include histograms of spatio-temporal filters and velocity distributions. This paper makes three contributions to the literature: i) Learning a dictionary of video primitives using parametric generative models; ii) Proposing the Spatio-Temporal FRAME (ST-FRAME) and Motion-Appearance FRAME (MA-FRAME) models for modeling and synthesizing textured motion; and iii) Developing a parsimonious hybrid model for generic video representation. Given an input video, VPS selects  the proper models automatically for different motion patterns and is compatible with high-level action representations. In the experiments, we synthesize a number of  textured motion; reconstruct real videos using the VPS; report a series of human perception experiments to verify the quality of reconstructed videos; demonstrate how the VPS changes over the scale transition in videos; and present the close connection between VPS and high-level action models.
\end{abstract}

\section{Introduction}
\label{intro}

\subsection{Motivation}

\begin{figure}
\begin{center}
   \includegraphics[width=1\linewidth]{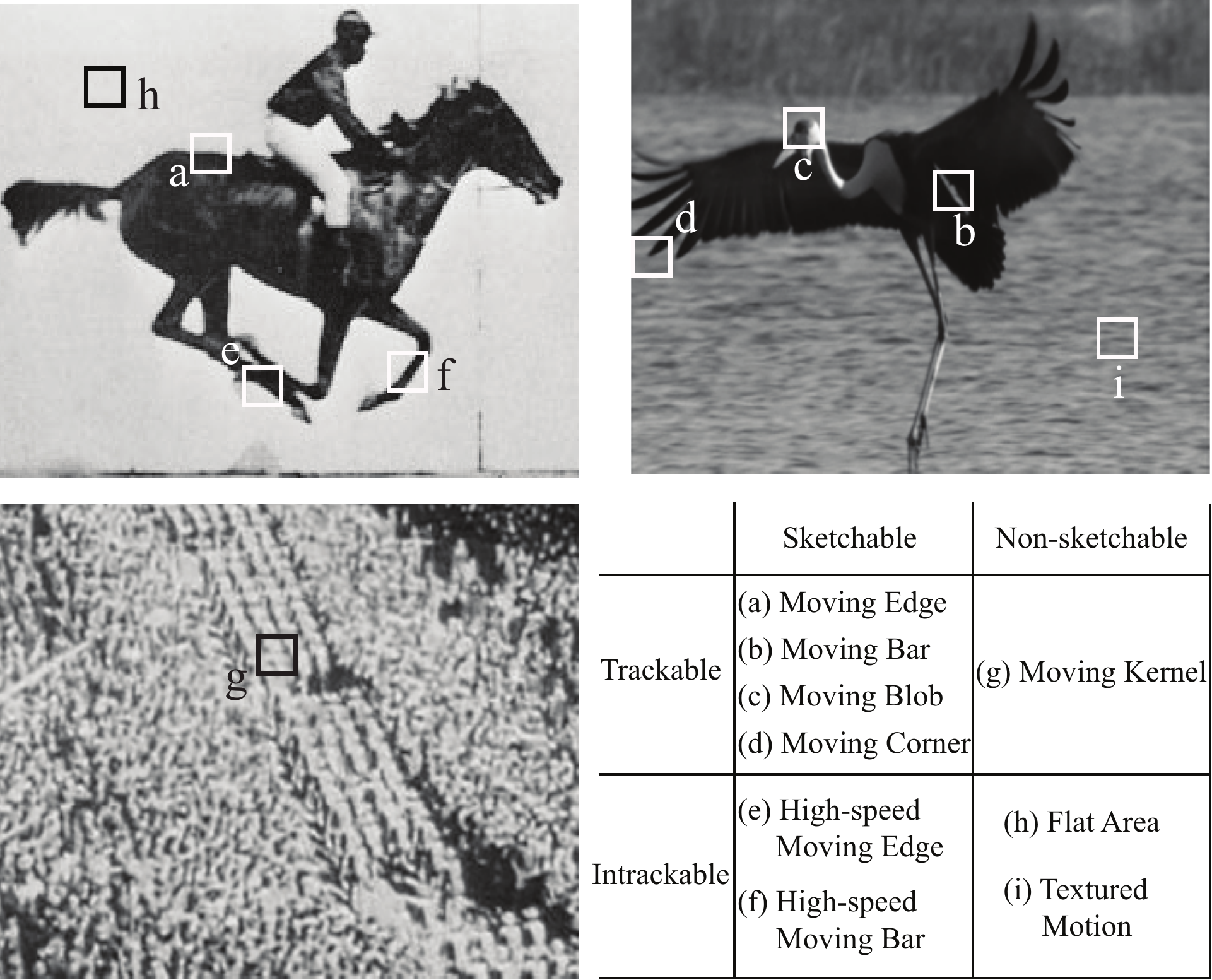}
\end{center}
   \caption{The four types of local video patches characterized by two criteria \--- sketchability and trackability.
   }
\label{examples}
\end{figure}

Videos of natural scenes contain vast varieties of motion patterns. We divide these motion patterns in a $2\times2$ table based on their complexities measured by two criteria: i) sketchability (\cite{Guo2007}), i.e. whether a local patch can be represented explicitly by an image primitive from a sparse coding dictionary, and ii) intrackability (or trackability) (\cite{Gong2010}), which measures the uncertainty of tracking an image patch using the entropy of posterior probability on velocities. Fig.\ref{examples} shows some examples of the different video patches in the four categories. Category A consists of the simplest vision phenomena, i.e. sketchable and trackable motions, such as trackable corners, lines, and feature points, whose positions and shapes can be tracked between frames. For example, patches (a), (b), (c) and (d) belong to category A. Category D is the most complex and is called textured motions or dynamic texture in the literature, such as water, fire or grass, in which the images have no distinct primitives or trackable motion, such as patches (h) and (i). The other categories are in between.  Category B refers to  sketchable but intrackable patches, which can be described by distinct image primitives but hardly be tracked between frames due to fast motion, for example the patches (e) and (f) at the legs of the galloping horse. Finally category C includes the trackable but non-sketchable patches, which are  cluttered features or moving kernels, e.g. patch (g).

In the vision literature, as it was pointed out by (\cite{Shi2007}), there are two families of  representations, which code images or videos by explicit and implicit functions respectively.

 1, Explicit representations with generative models. (\cite{Olshausen2003,Kim2010}) learned  an over-complete set of coding elements from natural video sequences  using the sparse coding model (\cite{Olshausen1996}). (\cite{Elder1998}) and (\cite{Guo2007})  represented the image/video patches by fitting  functions with explicit geometric and photometric parameters.  (\cite{Wang2004}) synthesized  complex motion, such as birds, snowflakes, and waves with  a large mount of particles and wave components. (\cite{Black2000}) represented two types of  motion primitives, namely smooth motion and motion boundaries for motion segmentation. In higher level object motion tracking, people represented different tracking units depending on the underlying objects and scales, such as  sparse or dense feature points tracking (\cite{Serby2004},\cite{Black2000}), kernels tracking (\cite{Comaniciu2003,Fan2006}),  contours tracking (\cite{Maccormick2000}), and middle-level pairwise-components generation (\cite{Yuan2010}).

 2, Implicit representations with descriptive models. For textured motions or dynamic textures, people used numerous Markov models which are constrained to reproduce some statistics extracted from the input video. For example, dynamic textures (\cite{Szummer1996,Campbell2002}) were modeled by a spatio-temporal auto-regressive (STAR) model, in which the intensity of each pixel was represented by a linear summation of intensities of its spatial and temporal neighbors.   (\cite{Bouthemy2006}) proposed a mixed-state auto-models for motion textures by generalizing the auto-models in (\cite{Besag1974}).  (\cite{Doretto2003}) derived an auto-regression moving-average  model for dynamic texture. (\cite{Chan2008}) and (\cite{Ravichandran2009}) extended it to a stable linear dynamical system (LDS) model.

Recently, to represent complex motion, such as human activities, researchers have used Histogram of Oriented Gradients (HOG) (\cite{Dalal2005}) for appearance and Histogram of Oriented Optical-Flow (HOOF) (\cite{Dalal2006,Chaudhry2009}) for motion. The HOG and HOOF record the rough geometric information through the grids and pool the statistics (histograms) within the local cells. Such features are used for recognition in discriminative tasks,  such as action classification, and are not suitable for video coding and reconstruction.

In the literature, these video representations are often manually selected for specific videos in different tasks. There lacks a generic representation and criterion that can automatically select the proper models for different patterns of the video. Furthermore, as it was demonstrated in (\cite{Gong2010}) that both sketchability and trackability change over \textit{scales}, \textit{densities}, and \textit{stochasticity of the dynamics},  a good video representation must adapt itself  continuously in a long video sequence.

\subsection{Overview and contributions}

Motivated by the above observations, we study a unified middle-level representation, called \textit{video primal sketch} (VPS), by integrating the two families of representations. Our work is inspired by Marr's conjecture  for a generic ``token'' representation called primal sketch as the output of early vision (\cite{Marr1982}), and is aimed at extending  the primal sketch model proposed by (\cite{Guo2007}) from images to videos.
Our goal is not only to provide a parsimonious model for video compression and coding, but more importantly, to support and be compatible with high-level tasks such as motion tracking and action recognition.

\begin{figure}
\begin{center}
   \includegraphics[width=1\linewidth]{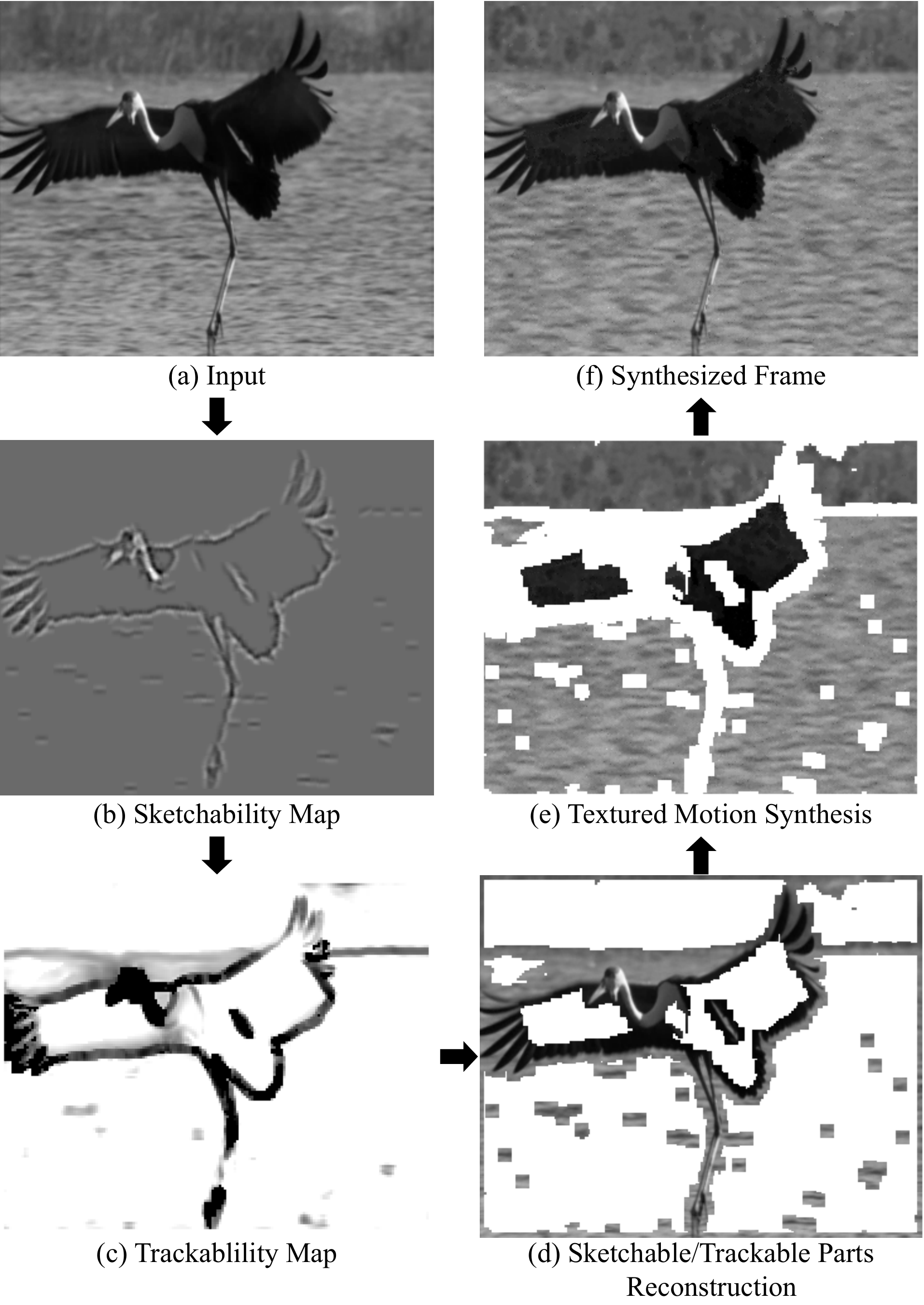}
\end{center}
   \caption{An example of Video Primal Sketch. (a) An input frame. (b)
   Sketchability map where dark means sketchable.
   (c) Trackability map where darker means higher trackability.
   (d) Reconstruction of explicit regions using primitives.
    (e) Synthesis for   implicit regions (textured motions) by sampling the generalize FRAME model through Markov chain Monte Carlo using the explicit regions as boundary condition. (f) Synthesized frame by
   combining the explicit and implicit representations.
   } \label{illustration}
\end{figure}

Fig.\ref{illustration} overviews an example of the video primal sketch. Fig.\ref{illustration}.(a) is an input video frame which is separated into sketchable and non-sketchable regions by the sketchability map in (b), and trackable primitives and intrackable regions by the trackability map in (c). The sketchable or trackable regions are explicitly represented by a sparse coding model and reconstructed in (d) with motion primitives, and each non-sketchable and intrackable region has a textured motion which is synthesized in (e) by a generalized FRAME (\cite{Zhu1998}) model (implicit and descriptive). The synthesis  of this frame is shown in (f) which integrates the results from (d) and (e) seamlessly.

As  Table~\ref{parameters} shows, the explicit representations include $3,600$ parameters for the positions, types, motion velocities, \textit{etc} of the video primitives and the implicit representations have $420$ parameters for the histograms of a set of filter responses on dynamic textures. This table shows the efficiency of the VPS model.

This paper makes the following contributions to the literature.

\begin{itemize}
\item[1.] We present and compare two different but related models to define textured motions. The first one is a spatio-temporal FRAME (ST-FRAME) model, which is a non-parametric Markov random field and generalizes the FRAME model (\cite{Zhu1998}) of texture with spatio-temporal filters. The ST-FRAME model is learned so that it has marginal probabilities that match the histograms of the responses from the spatio-temporal filters on the input video.  The second one is a motion-appearance FRAME model (MA-FRAME), which not only matches the histograms of some spatio-temporal filter responses, but also matches the histograms of  velocities pooled over a local region.  The MA-FRAME  model achieves better results in video synthesis than the ST-FRAME model, and it is, to some extent, similar to the HOOF features used in action classification (\cite{Dalal2006,Chaudhry2009}).

\item[2.] We learn a  dictionary of motion primitives from input videos using a generative sparse coding  model. These primitives are used to reconstruct the explicit regions and include two types: i) generic primitives for the sketchable patches, such as corners, bars etc; and ii) specific primitives for the non-sketchable but trackable patches which are usually texture patches similar to those used in
 kernel tracking (\cite{Comaniciu2003}).

\item[3.] The models for implicit and explicit regions are integrated in a hybrid representation \--- the video primal sketch (VPS), as a generic middle-level representation of video. We will also show how VPS changes over information scales affected by distance, density and dynamics.

\item[4.]  We show the connections between this middle-level VPS  representation and  features for high-level vision tasks such as action recognition.
\end{itemize}

\begin{table}
\caption{The parameters in video primal sketch model for the water
bird video in Fig.\ref{illustration}}\label{parameters}
\begin{center}
\begin{tabular}{|c|c|}
\hline
Video Resolution & 288$\times$352 pixels \\
\hline
Explicit Region & 31,644 pixels$\approx$ 30\% \\
\hline
Primitive Number & 300 \\
\hline
Primitive Width & 11 pixels\\
\hline
Explicit Parameters & 3,600 $\approx$ 3.6\% \\
\hline
Implicit parameters & 420 \\
\hline
\end{tabular}
\end{center}
\end{table}

Our work is inspired by Gong's empirical study in \cite{Gong2010}, which revealed the statistical properties of videos over scale transitions and defined intrackability as the entropy of local velocities. When the entropy is high, the patch cannot be tracked locally and thus its motion is represented by a velocity histogram.  \cite{Gong2010} did not give a unified model for video representation and synthesis which is the focus on the current paper.

This paper extends a previous conference paper (\cite{Han2011}) in the following aspects:

\begin{itemize}
\item[1.] We propose a new dynamic texture model, MA-FRAME, for better representing velocity information. Benefited from the new temporal feature, the VPS model can be applied to high-level action representation tasks more directly.

\item[2.] We compare spatial and temporal features with HOG (\cite{Dalal2005}) and HOOF (\cite{Dalal2006}) and discuss the connections between them.

\item[3.] We do a series of perceptual experiments to verify the high quality of video synthesis from the aspect of human perception.
\end{itemize}

The remainder of this paper is organized as follows. In Section 2, we present the  framework of video primal sketch. In Section 3, we explain the algorithms for explicit representation, textured motion synthesis and video synthesis, and show a series of experiments. The paper is concluded with a discussion in Section 4.

\section{Video primal sketch model}

In his monumental book (\cite{Marr1982}),  Marr conjectured a primal sketch as the output of early vision that transfers the continuous ``analogy'' signals in pixels to a discrete ``token'' representation. The latter should be parsimonious and sufficient to reconstruct the observed image without much perceivable distortions. A mathematical model was later studied by Guo, \textit{et al} (\cite{Guo2007}), which successfully modeled hundreds of images by integrating sketchable structures and non-sketchable textures. In this section, we extend it to video primal sketch as a hybrid generic video representation.

Let $\mathbf{I}[1,m]=\{\mathbf{I}_{(t)}\}_{t=1}^m$ be a video defined on a 3D lattice $\Lambda\subset Z^3$. $\Lambda$ is divided disjointly into explicit and implicit regions,
\begin{gather}\label{parts}
\Lambda=\Lambda_{ex}\bigcup \Lambda_{im},\;\Lambda_{ex}\bigcap\Lambda_{im}=\emptyset.
\end{gather}
Then the video $\mathbf{I}$ is decomposed as two components
\begin{gather}\label{decompose}
\mathbf{I}_{\Lambda}=(\mathbf{I}_{\Lambda_{ex}},\mathbf{I}_{\Lambda_{im}}).
\end{gather}
$\mathbf{I}_{\Lambda_{ex}}$ are defined by explicit functions $I=g(w)$, in which, each instance is corresponded to a different function form of $g()$ and indexed by a particular value of parameter $w$. And $\mathbf{I}_{\Lambda_{im}}$ are defined by implicit functions $H(I)=h$, in which, $H()$ extracts the statistics of filter responses from image $I$ and $h$ is a specific value of histograms.

In the following, we first present the two families of models for $\mathbf{I}_{\Lambda_{ex}}$ and $\mathbf{I}_{\Lambda_{im}}$ respectively, and then integrate them in the VPS model.

\subsection{Explicit representation by sparse coding}\label{ExRep}

The explicit region $\Lambda_{ex}$ of a video $\mathbf{I}$ is decomposed into $n_{ex}$ disjoint domains (usually $n_{ex}$ is in the order of $10^2$),
\begin{gather}\label{primalsketch}
\Lambda_{ex}=\bigcup_{i=1}^{n_{ex}}\Lambda_{ex,i}.
\end{gather}
Here $\Lambda_{ex,i}\subset\Lambda$ defines the domain of a ``brick''. A brick, denoted by $\mathbf{I}_{\Lambda_{ex,i}}$, is a spatio-temporal volume like a patch in images.  These bricks are divided into the three categories A, B and C as we mentioned in section \ref{intro}.

The size of $\Lambda_{ex,i}$ influences the results of tracking and synthesis to some degree. The spatial size should depend on the scale of structures or the granularity of textures, and the temporal size should depend on the motion amplitude and frequency in time dimension, which are hard to estimate in real applications. However, a general size works well for most of cases, say $ 11\times11$ pixels$\times$ 3 frames for trackable bricks (sketchable or non-sketchable), or $11\times11$ pixels$\times$1 frame for sketchable but intrackable bricks. Therefore, in all the experiments of this paper, the size of $\Lambda_{ex,i}$ is chosen as such.

\begin{figure}
\begin{center}
   \includegraphics[width=1.0\linewidth]{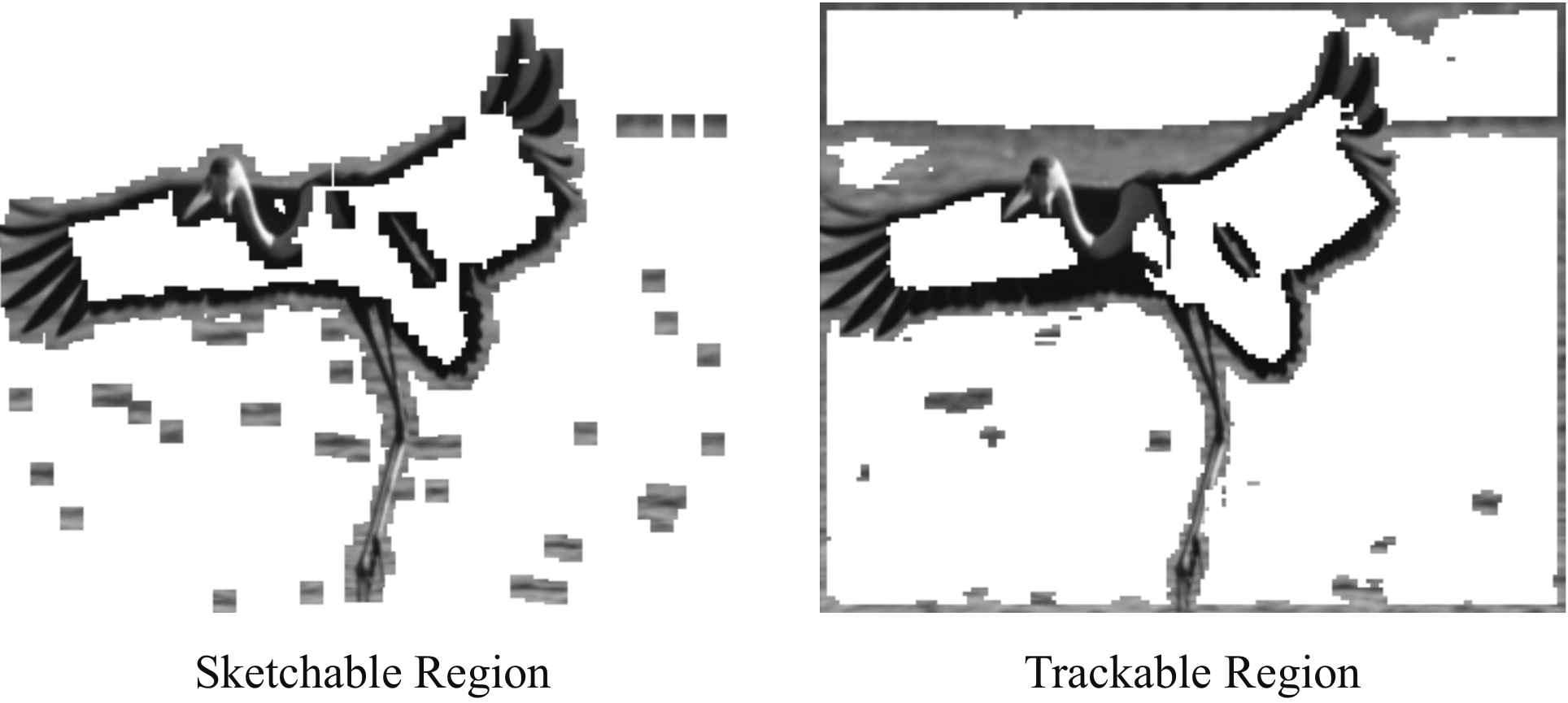}
\end{center}
   \caption{Comparison between sketchable and trackable regions.}
\label{sk_vs_tr}
\end{figure}

Fig.\ref{sk_vs_tr} shows one example comparing the sketchable and trackable regions based on sketchability and trackability maps shown in Fig.\ref{illustration}(b) and (c) respectively. It is worth noting that the two regions overlaps with only a small percentage of the regions is either sketchable or trackable.

Each brick can be represented by a primitive $B_i\in\Delta_B$ through an explicit function,
\begin{gather}\label{trackable}
\mathbf{I}(x,y,t)=\alpha_i B_i(x,y,t)+\epsilon,\quad\forall(x,y,t)\in\Lambda_{ex,i}.
\end{gather}
$B_i$ means the $i$th primitive from the primitive dictionary $\Delta_B$, which fits the brick $\mathbf{I}_{\Lambda_{ex,i}}$ best. Here $i$ indexes the parameters such as type, position, orientation and scale of $B_i$. $\alpha_i$ is the corresponding coefficient. $\epsilon$ represents the residue, which is assumed to be i.i.d. Gaussian.
For an trackable primitive, $B_i(x,y,t)$ includes $3$ frames and thus encodes the velocity $(u,v)$ in the $3$ frames. For sketchable but intrackable primitive, $B_i(x,y,t)$ has only $1$ frame.

\begin{figure}
\begin{center}
   \includegraphics[width=0.8\linewidth]{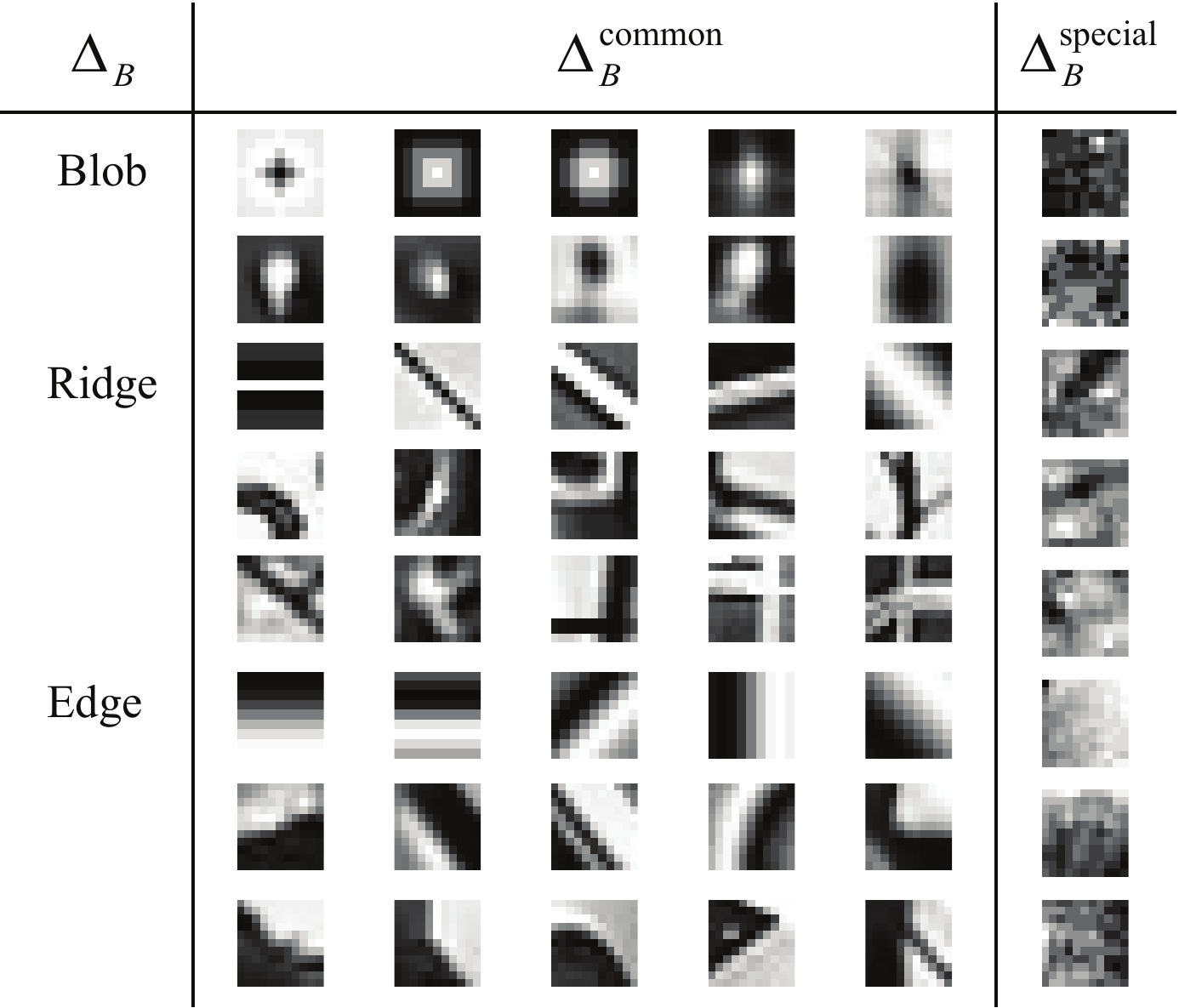}
\end{center}
   \caption{Some selected examples of primitives. $\Delta_B$ is a dictionary of primitives with velocities (u,v) ( (u,v) is not shown), such as blobs,
   ridges, edges and special primitives.}
\label{primitive_example}
\end{figure}

As Fig.~\ref{primitive_example} illustrates,  the dictionary $\Delta_B$ is composed of two categories:
\begin{itemize}
\item Common primitives $\Delta_B^{common}$. These are primitives shared by most videos, such as blobs, edges and ridges etc. They have explicit parameters for orientations and scales. They are mostly belong to sketchable region as shown in Fig. \ref{sk_vs_tr}.

\item Special primitives $\Delta_B^{special}$. These bricks  do not have common appearance and are limited to specific video frames. They are non-sketchable but trackable, and are recorded to code the specific video region. They are mostly belong to trackable region but not included in sketchable region as shown in Fig. \ref{sk_vs_tr}.
\end{itemize}
To be noted, the primitives and categories shown in Fig.~\ref{primitive_example} are some selected examples, but not the whole dictionary. The details for learning these primitives are introduced in section \ref{learning}.

(\ref{trackable}) uses only one base function and thus is different from conventional linear additive model. Follows the Gaussian assumption for the residues, we have the following probabilistic model for the explicit region $\mathbf{I}_{\Lambda_{ex}}$
\begin{gather}\label{generativemodel}
\begin{aligned}
&p(\mathbf{I}_{\Lambda_{ex}};\mathbf{B},\alpha)=\prod_{i=1}^{n_{ex}}\frac{1}{(2\pi)^{\frac{n}{2}}\sigma_i^n}\exp\{-E_i\}\\
&E_i=\sum_{(x,y,t)\in\Lambda_{ex,i}}\frac{(\mathbf{I}(x,y,t)-\alpha_iB_i(x,y,t))^2}{2\sigma_i^2}.
\end{aligned}
\end{gather}
where $\mathbf{B}=(B_1,...,B_{n_{ex}})$ represents the selected primitive set, $n$ is the size of each primitive, $n_{ex}$ is the number of selected primitives and $\sigma_i$ is estimated standard deviation of representing natural videos by $B_i$.

\subsection{Implicit representations by FRAME models}\label{ImRep}

The implicit region $\Lambda_{im}$ of video $\mathbf{I}$ is segmented into $n_{im}$ (usually $n_{im}$ is no more than $10$) disjoint homogeneous textured motion regions,
\begin{gather}\label{intrackable}
\Lambda_{im}=\bigcup_{j=1}^{n_{im}} \Lambda_{im,j}.
\end{gather}

One effective approach for texture modeling is to pool the histograms for a set of filters (Gabor, DoG and DooG) on the input image (\cite{Bergen1991, Chubb1991, Heeger1995, Zhu1998, Portilla2000}). Since Gabor filters model the response functions of the neurons in the primary visual cortex, two texture images with the same histograms of filter responses generate the same texture impression, and thus are considered perceptually equivalent (\cite{Silverman1989}). The FRAME model proposed in (\cite{Zhu1998}) generates the expected marginal statistics to match the observed histograms through the maximum entropy principle. As a result, any images drawn from this model will have the same filtered histograms and thus can be used for synthesis or reconstruction.

We extend this concept to video by adding temporal constraints and define each homogeneous textured motion region $\mathbf{I}_{\Lambda_{im,j}}$ by  an equivalence class of videos,
\begin{gather}\label{ensemble}
\Omega_K(\mathbf{h}_j)=\{\mathbf{I}_{\Lambda_{im,j}}:
H_k(\mathbf{I}_{\Lambda_{im,j}})= h_{k,j}, k=1,2,...,K\}.
\end{gather}
where $\mathbf{h}_j=(h_{1,j},...,h_{K,j})$ is a series of 1D histograms of filtered responses that characterize the macroscopic properties of the textured motion pattern.   Thus we only need to code the histograms $\mathbf{h}_j$ and synthesize the textured motion region $\mathbf{I}_{\Lambda_{im,j}}$  by sampling from the set $\Omega_K(\mathbf{h}_j)$.  As $\mathbf{I}_{\Lambda_{im,j}}$ is defined by the implicit  functions, we call it an implicit representation.
These regions are coded up to an equivalence class in contrast to reconstructing the pixel intensities in the explicit representation.

To capture temporal constraints, one straightforward method is to choose a set of spatio-temporal filters and calculate the histograms of the filter responses.  This leads to the spatio-temporal FRAME (ST-FRAME) model which will be introduced in section \ref{DT}.
Another method is to compute the statistics of velocity. Since the motion in these regions is intrackable, at each point of the image, its velocity is ambiguous (large entropy). We pool the histograms of velocities locally in a way similar to the  HOOF (Histogram of Oriented Optical-Flow)(\cite{Dalal2006,Chaudhry2009}) features in action classification. This leads to the motion-appearance FRAME (MA-FRAME) model which uses histograms of both appearance (static filters) and velocities.  We will elaborate on this model in section \ref{VF}.

\subsection{Implicit representation by spatio-temporal FRAME}\label{DT}

ST-FRAME is an extension of the FRAME model (\cite{Zhu1998}) by adopting spatio-temporal filters.

\begin{figure}
\begin{center}
   \includegraphics[width=0.8\linewidth]{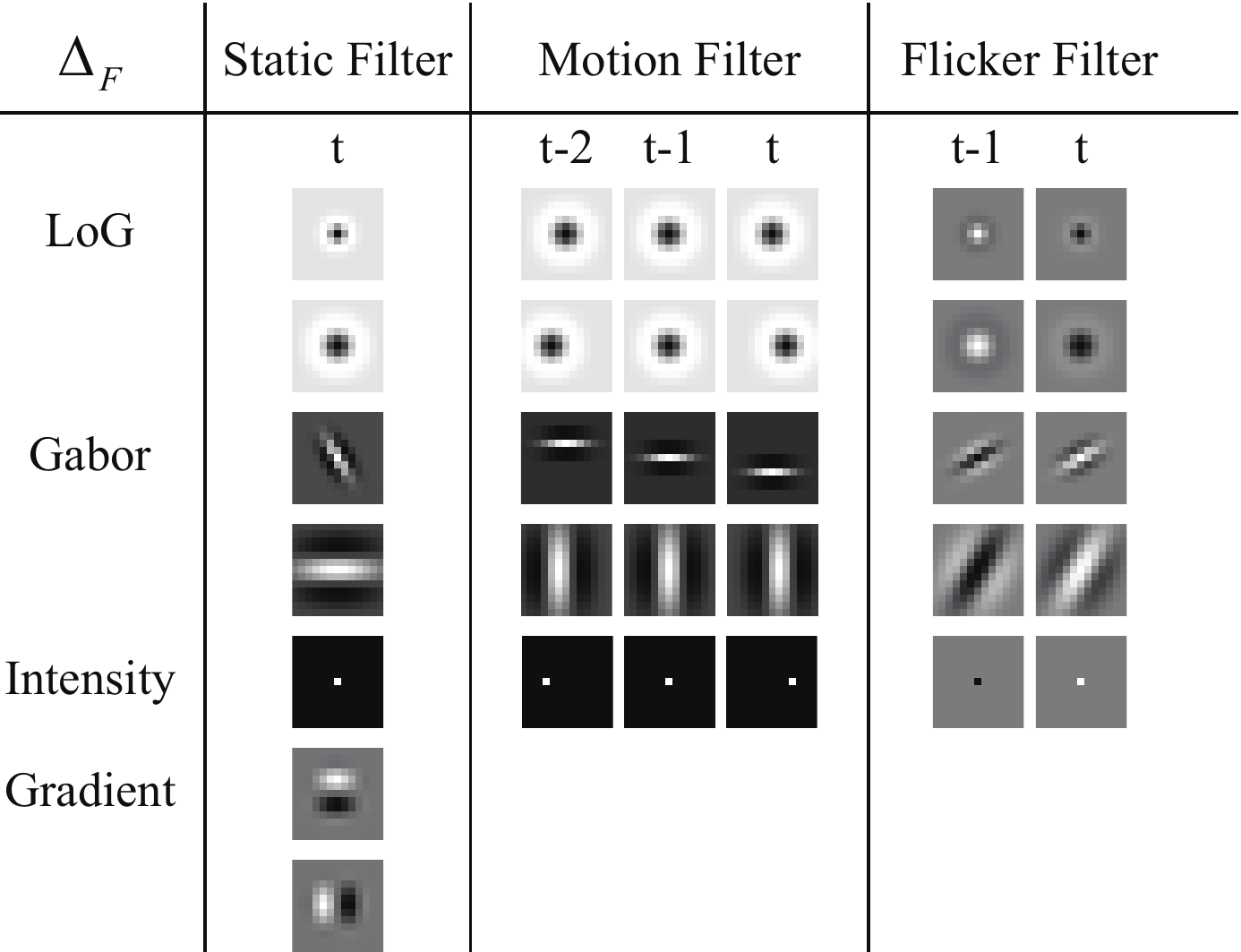}
\end{center}
   \caption{$\Delta_F$ is a dictionary of spatio-temporal filters including static, motion and flicker filters.}
\label{filter_example}
\end{figure}

A set of filters $\mathbf{F}$ is selected from a filter bank $\Delta_F$. Fig.\ref{filter_example} illustrates the three types of filters in $\Delta_F$: i) the static filters for texture appearance in a single image; ii) the motion filter with certain velocity; and iii) the flicker filter that have zero velocity but opposite signs between adjacent frames.  For each filter $F_k\in\mathbf{F}$, the spatio-temporal filter response of $\mathbf{I}$ at $(x,y,t)\in\Lambda_{im,j}$ is $F_k*\mathbf{I}(x,y,t)$.  The convolution is over spatial and temporal domain. By pooling the filter responses over all $(x,y,t)\in \Lambda_{im,j}$, we obtain a number of 1D histograms
\begin{align}
&H_k(\mathbf{I}_{\Lambda_{im,j}})=H_k(z;\mathbf{I}_{\Lambda_{im,j}})\label{hist}\\
&=\frac{1}{|\Lambda_{im,j}|}\sum_{(x,y,t)\in\Lambda_{im,j}}{\delta(z;F_k*\mathbf{I}(x,y,t))},k=1,...,K.\notag
\end{align}
where $z$ indexes the histogram bins, and $\delta(z;x)=1$ if $x$ belongs to bin $z$, and $\delta(z;x)=0$ otherwise. Following the FRAME model, the statistical model of textured motion $\mathbf{I}_{\Lambda_{im,j}}$ is written in the form of the following Gibbs distribution,
\begin{gather}\label{descriptivemodel}
p(\mathbf{I}_{\Lambda_{im,j}};\mathbf{F},\beta)\propto\exp\{-\sum_k\langle\beta_{k,j},H_k(\mathbf{I}_{\Lambda_{im,j}})\rangle\}.
\end{gather}
where $\beta_k=(\beta_{k,1},\beta_{k,2},...\beta_{k,3})$ are potential functions.

According to the theorem of ensemble equivalence~(\cite{Wu2000}), the Gibbs distribution converges to the uniform distribution over the set $\Omega_K(\mathbf{h}_j)$ in (\ref{ensemble}), when $\Lambda_{im,j}$ is large enough. For any fixed local brick $\Lambda_0\subset\Lambda_{im,j}$, the distribution of $I_{\Lambda_0}$ follows the Markov random field model (\ref{descriptivemodel}). The model can describe textured motion located in an irregular shape region $\Lambda_{im,j}$.

The filters in $\mathbf{F}$ are pursued one by one from the filter bank $\Delta_F$ so that the information gain is maximized at each step.
\begin{gather}\label{infogain}
F_k^*=\arg\max_{F_k\in\Delta_F}\Vert H_k^{syn}-H_k^0\Vert.
\end{gather}
$H_k^0$ and $H_k^{syn}$ are the response histograms of $F_k$ before and after synthesizing $\mathbf{I}_{\Lambda_{im,j}}$ by adding $F_k$ respectively. The larger the difference, the more important is the filter.

Following the distribution form of (\ref{descriptivemodel}), the probabilistic model of implicit parts of $\mathbf{I}$ is defined as
\begin{gather}\label{intrackablemodel}
p(\mathbf{I}_{\Lambda_{im}};\mathbf{F},\beta)\propto\prod_{j=1}^{n_{im}}p(\mathbf{I}_{\Lambda_{im,j}};\mathbf{F},\beta).
\end{gather}
where $\mathbf{F}=(F_1,...,F_K)$ represents the selected spatio-temporal filter set.

In the experiments described later, we demonstrate that this model can synthesize a range of dynamic textures by matching the histograms of filter responses. The synthesis is done through sampling the probability by Markov chain Monte Carlo.

\subsection{Implicit representation by motion-appearance FRAME}\label{VF}

Different from ST-FRAME, in which, temporal constraints are based on  spatio-temporal filters, the MA-FRAME model uses the statistics of velocities, in addition to the statistics of filter responses for appearance.

For the appearance constraints, the filter response histograms $H^{(s)}$ are obtained similarly as ST-FRAME in (\ref{hist})
\begin{align}
&H^{(s)}_k(\mathbf{I}_{\Lambda_{im,j}})=H^{(s)}_k(z;\mathbf{I}_{\Lambda_{im,j}})\label{hist_s}\\
&=\frac{1}{|\Lambda_{im,j}|}\sum_{(x,y,t)\in\Lambda_{im,j}}{\delta(z;F_k*\mathbf{I}(x,y,t))},k=1,...,K.\notag
\end{align}
where the filter set $\mathbf{F}$ includes static and flicker filters in $\Delta_F$.

For the motion constraints, the velocity distribution of each local patch is estimated via the calculation of trackability (\cite{Gong2010}), in which, each patch is compared with its spatial neighborhood in adjacent frame and the probability of the local velocity $v$ is computed as
\begin{align}
&p(v|I(x,y,t-1),I(x,y,t))\label{ssd}\\
&\propto\exp\{-\frac{\Vert I_{\partial(x-v_x,y-v_y,t)}-I_{\partial(x,y,t-1)}\Vert^2}{2\sigma^2}\}.\notag
\end{align}
Here, $\sigma$ is the standard deviation of the differences between local patches from adjacent frames based on various velocities. The statistical information of velocities for a certain area of texture is approximated by averaging the velocity distribution over region  $\Lambda_{im,j}$
\begin{align}
&H^{(t)}(\mathbf{v}_{\Lambda_{im,j}})\label{hist_v}\\
&=\sum_{(x,y,t)\in\Lambda_{im,j}}{p(v|I(x,y,t-1),I(x,y,t))}.\notag
\end{align}


Let $\mathbf{H}=( \mathbf{H}^{(s)}(\mathbf{I}_{\Lambda_{im,j}}),\mathbf{H}^{(t)}(\mathbf{v}_{\Lambda_{im,j}}) )$ collect the filter responses and velocities histograms of the video.
The statistical model of textured motion
$\mathbf{I}_{\Lambda_{im,j}}$ can be written in the form of the
following joint Gibbs distribution,
\begin{gather}\label{descriptivemodel2}
p(\mathbf{I}_{\Lambda_{im,j}}; \mathbf{F},\beta)
\propto\exp\{-\langle\mathbf{\beta},\mathbf{H}(\mathbf{I}_{\Lambda_{im,j}},\mathbf{v}_{\Lambda_{im,j}})\rangle\}.
\end{gather}
Here, $\beta$ is the parameter of the model.

In summary, the probabilistic model for the implicit regions of $\mathbf{I}$ is defined as
\begin{gather}\label{intrackablemodel2}
p(\mathbf{I}_{\Lambda_{im}};\mathbf{F},\beta)\propto\prod_{j=1}^{n_{im}}p(\mathbf{I}_{\Lambda_{im,j}};\mathbf{F},\beta).
\end{gather}
where $\mathbf{F}=(F_1,...,F_K)$ represents the selected filter set.

In the experiment section, we show the effectiveness of the MA-FRAME model and its advantages over the ST-FRAME model.

\subsection{Hybrid model for video representation}

The ST or MA-FRAME models for the implicit regions $\mathbf{I}_{\Lambda_{im}}$ use the  explicit regions $\mathbf{I}_{\Lambda_{ex}}$ as  boundary conditions, and the probabilistic models for
$\mathbf{I}_{\Lambda_{ex}}$ and $\mathbf{I}_{\Lambda_{im}}$ are
given by (\ref{generativemodel}) and (\ref{intrackablemodel}) respectively
\begin{gather}\label{structuralpart}
\mathbf{I}_{\Lambda_{ex}}\sim
p(\mathbf{I}_{\Lambda_{ex}};\mathbf{B},\alpha),\mathbf{I}_{\Lambda_{im}}\sim
p(\mathbf{I}_{\Lambda_{im}}|\mathbf{I}_{\partial\Lambda_{im}};\mathbf{F},\beta).
\end{gather}
Here, $\mathbf{I}_{\partial\Lambda_{im}}$ represents the boundary condition of $\mathbf{I}_{\Lambda_{im}}$, which belong to the reconstruction of $\mathbf{I}_{\Lambda_{ex}}$. It leads to seamless
boundaries in the synthesis.

By integrating the explicit and implicit representation,  the video primal sketch has the following probability model,
\begin{align}
&p(\mathbf{I}|\mathbf{B},\mathbf{F},\alpha,\beta)=\notag\\&\frac{1}{Z}\exp\{-\sum_{i=1}^{n_{ex}}\sum_{(x,y,t)\in
\Lambda_{ex,i}}\frac{(\mathbf{I}(x,y,t)-\alpha_iB_i(x,y,t))^2}{2\sigma_i^2}
\notag\\&-\sum_{j=1}^{n_{im}}\sum_{k=1}^K\langle\beta_{k,j},H_k(\mathbf{I}_{\Lambda_{im,j}}|\mathbf{I}_{\partial\Lambda_{im,j}})\rangle\}.\label{VPS}
\end{align}
where $Z$ is the normalizing constant.

We denote by $VPS=(\mathbf{B},\mathbf{H})$  the representation for the video $\mathbf{I}_{\Lambda}$, where
$\mathbf{H}=(\{h_{k,1}\}_{k=1}^{K_1},...,\{h_{k,n_{im}}\}_{k=1}^{K_{n_{im}}})$ includes the histograms described by $\mathbf{F}$ and $\mathbf{V}$; and $\mathbf{B}$ includes all the primitives with parameters for their indexes, position, orientation and scales etc. $p(VPS)=p(\mathbf{B},\mathbf{H})=p(\mathbf{B})p(\mathbf{H})$ gives the prior probability of video representation by $VPS$. $p(\mathbf{B})\propto\exp\{-|\mathbf{B}|\}$, in which, $|\mathbf{B}|$ is the number of primitives. $p(\mathbf{H})\propto\exp\{-\gamma_{tex}(\mathbf{H})\}$, in which, $\gamma_{tex}(\mathbf{H})$ is the energy term and for instance, $\gamma_{tex}(\mathbf{H})=\rho n_{im}$ to penalize the number of implicit regions. Thus, the best video representation $VPS^*$ is obtained by maximizing the posterior probability,
\begin{align}
&VPS^*=\arg\max_{VPS} p(VPS|\mathbf{I}_{\Lambda})=\arg\max_{VPS} p(\mathbf{I}_{\Lambda}|VPS)p(VPS)\notag\\
&=\arg\max_{\mathbf{B},\mathbf{H}}\frac{1}{Z}\exp\{-\sum_{i=1}^{n_{ex}}\sum_{(x,y,t)\in \Lambda_{ex,i}}\frac{(\mathbf{I}(x,y,t)-\alpha_iB_i(x,y,t))^2}{2\sigma_i^2}\notag\\
&-|\mathbf{B}|-\sum_{j=1}^{n_{im}}\sum_{k=1}^K\langle\beta_{k,j},H_k\rangle-\gamma_{tex}(\mathbf{H})\}.\label{V}
\end{align}
following the video primal sketch model in (\ref{VPS}).

Table~\ref{parameters} shows an example of $VPS$. For a video of the size of $288\times352$ pixels, about 30\% of the pixels are represented explicitly by $n_{ex}=300$ motion primitives. As each primitive needs 11 parameters (the side length of the patch according to the primitive learning process in section \ref{learning}) to record the profile and 1 more to record the type, the number of total parameters for the explicit representation is 3,600. $n_{im}=3$ textured motion regions are represented implicitly by the histograms, which are described by $K_1=11$, $K_2=12$ and $K_3=5$ filters respectively. As each histogram has 15 bins, the number of the parameters for the implicit representation is 420.

\subsection{Sketchability and Trackability for Model Selection}\label{modelselect}

The computation of the VPS involves the  partition of the domain $\Lambda$ into the explicit regions $\Lambda_{ex}$ and implicit regions $\Lambda_{im}$. This is done through the sketchability and trackability maps. In this subsection, we overview the general ideas and refer to previous work on sketchability (\cite{Guo2007}) and trackability (\cite{Gong2010}) for details.

Let's consider one local volume $\Lambda_0\subset\Lambda$ of the video $\mathbf{I}$. In the video primal sketch model, $\mathbf{I}_{\Lambda_0}$ may be modeled either by the sparse coding model in (\ref{generativemodel}) or by the FRAME model in (\ref{intrackablemodel}). The choice is determined via the competition between the two models, i.e. comparing which model gives shorter coding length (\cite{Shi2007}) for representation.

If $\mathbf{I}_{\Lambda_0}$ is represented by the sparse coding model, the posterior probability is calculated by
\begin{gather}
p(\mathbf{B}|\mathbf{I}_{\Lambda_0})=\frac{1}{(2\pi)^{n/2}\sigma^n}\exp\{-\sum_i\frac{\Vert\mathbf{I}_{\Lambda_0}-\alpha_i B_i\Vert^2}{2\sigma^2}\}.\label{P2}
\end{gather}
where $n=|\Lambda_0|$. The coding length is
\begin{gather}
L_{ex}(\mathbf{I}_{\Lambda_0})=\log\frac{1}{p(\mathbf{B}|\mathbf{I}_{\Lambda_0})}=\frac{n}{2}\log2\pi\sigma^2+\sum_i\frac{\Vert\mathbf{I}_{\Lambda_0}-\alpha_i B_i\Vert^2}{2\sigma^2}.\notag
\end{gather}
Since $\sigma^2$ is estimated via the given data temporarily in real application, $\frac{1}{n}\sum_i \Vert\mathbf{I}_{\Lambda_0}-\alpha_i B_i\Vert^2=\sigma^2$ holds by definition.
As a result, the coding length is derived as,
\begin{gather}
L_{ex}(\mathbf{I}_{\Lambda_0})=\frac{n}{2}(\log2\pi\sigma^2+1).\label{IC1}
\end{gather}

If $\mathbf{I}_{\Lambda_0}$ is described by the FRAME model, the posterior probability is calculated by
\begin{gather}\label{P1}
p(\mathbf{F}|\mathbf{I}_{\Lambda_0})\propto\exp\{-\sum_{k=1}^K\langle\beta_{k},H_k(\mathbf{I}_{\Lambda_0})\rangle\}.
\end{gather}
The coding length is estimated through a sequential reduction process. When $K=0$, with no constraints, the FRAME model is a uniform distribution, and thus the coding length is $\log|\Omega_0|$ where $|\Omega_0|$ is the cardinality of the space of all videos in $\Lambda$.
Suppose  the intensities of the video range from 0 to 255, then
$\log|\Omega_0| = 8 \times |\Lambda_0|$. By adding each constraint, the
 equivalence  $\Omega(K)$ will shrink in size, and the ratio of the compression $\log\frac{|\Omega_{K-1}|}{|\Omega_K|}$ is approximately equal to the information gain in (\ref{infogain}).
Therefore we can calculate the coding length by
\begin{gather}\label{E1}
L_{im}(\mathbf{I}_{\Lambda_0})=\log|\Omega_0|-\log\frac{|\Omega_0|}{|\Omega_1|}-...-\log\frac{|\Omega_{K-1}|}{|\Omega_K|}.
\end{gather}
By comparing $L_{im}(\mathbf{I}_{\Lambda_0})$ and $L_{ex}(\mathbf{I}_{\Lambda_0})$, whoever has the shorter coding length will win the competition and be chosen for $\mathbf{I}_{\Lambda_0}$.

In practice, we use a faster estimation which utilizes the relationship between the coding length and the entropy of the local posterior probabilities.

Consider the entropy of $p(B|\mathbf{I}_{\Lambda_0})$,
\begin{gather}\label{H2}
\mathcal{H}(B|\mathbf{I}_{\Lambda_0})=-E_{p(B,\mathbf{I}_{\Lambda_0})}[\log p(B|\mathbf{I}_{\Lambda_0})].
\end{gather}
It measures the uncertainty of selecting a primitive in $\Delta_B$ for representation.  The sharper the distribution $p(B|\mathbf{I}_{\Lambda_0})$ is, the lower the entropy $\mathcal{H}(B_k|\mathbf{I}_{\Lambda_0})$ will be, which gives smaller $L_{ex}(\mathbf{I}_{\Lambda_0})$ according to (\ref{IC1}).  Hence, $\mathcal{H}(B_k|\mathbf{I}_{\Lambda_0})$ reflects the magnitude of $L_\text{diff}(\mathbf{I}_{\Lambda_0})=L_{ex}(\mathbf{I}_{\Lambda_0})-L_{im}(\mathbf{I}_{\Lambda_0})$. Set an entropy threshold $\mathcal{H}_0$ on $\mathcal{H}(B_k|\mathbf{I}_{\Lambda_0})$, ideally, $\mathcal{H}(B_k|\mathbf{I}_{\Lambda_0})=\mathcal{H}_0$ if and only if $L_\text{diff}(\mathbf{I}_{\Lambda_0})=0$. Therefore, when $\mathcal{H}(B_k|\mathbf{I}_{\Lambda_0})<H_0$, we consider $L_{ex}(\mathbf{I}_{\Lambda_0})$ is lower and $\mathbf{I}_{\Lambda_0}$ is modeled by the sparse coding model, else it is modeled by the FRAME model.

It is clear that $\mathcal{H}(B_k|\mathbf{I}_{\Lambda_0})$ has the same form and meaning with sketchability (\cite{Guo2007}) in appearance representation and trackability (\cite{Gong2010}) in motion representation. Therefore, sketchability and trackability can be used for model selection for each local volume. Fig.\ref{illustration} (b) and (c) show the sketchability and trackability maps calculated by the local enetropy of posteriors. The two maps decide the partition of the video into the explicit implicit regions. Within the explicitly regions, they also decide whether a patch is trackable (using primitives with size of $11\times 11$ pixels $\times 3$ frames) or intrackable (using primitives with $11\times 11$ pixels $\times 1$ frame).

\section{Algorithms and experiments}

\subsection{Spatio-temporal filters}\label{MF}

In the vision literature, spatio-temporal filters have been widely used for motion information extraction (\cite{Adelson1985}), optical flow estimation (\cite{Heeger1987}), multi-scale representation of temporal data (\cite{Lindeberg1996}), pattern categorization (\cite{Wildes2000}), and dynamic texture recognition (\cite{Derpanis2010}).
In the experiments, we choose spatio-temporal filters $\Delta_F$ as shown in Fig.\ref{filter_example}. It includes three types:

\hspace{0.1cm}

\begin{itemize}
\item[1]  \textbf{Static filters.} Laplacian of Gaussian (LoG), Gabor, gradient, or intensity filter on
a single frame. They capture statistics of spatial features.

\item[2]  \textbf{Motion filters.} Moving LoG, Gabor or intensity filters in different speeds and
directions over three frames.  Gabor motion filters
move perpendicularly to their orientations.

\item[3]  \textbf{Flicker filters.} One static filter with opposite signs at two frames. They contrast
the static filter responses between two consequent frames and detect
the change of dynamics.
\end{itemize}

\hspace{0.1cm}

\begin{figure*}
\begin{center}
   \includegraphics[width=1\linewidth]{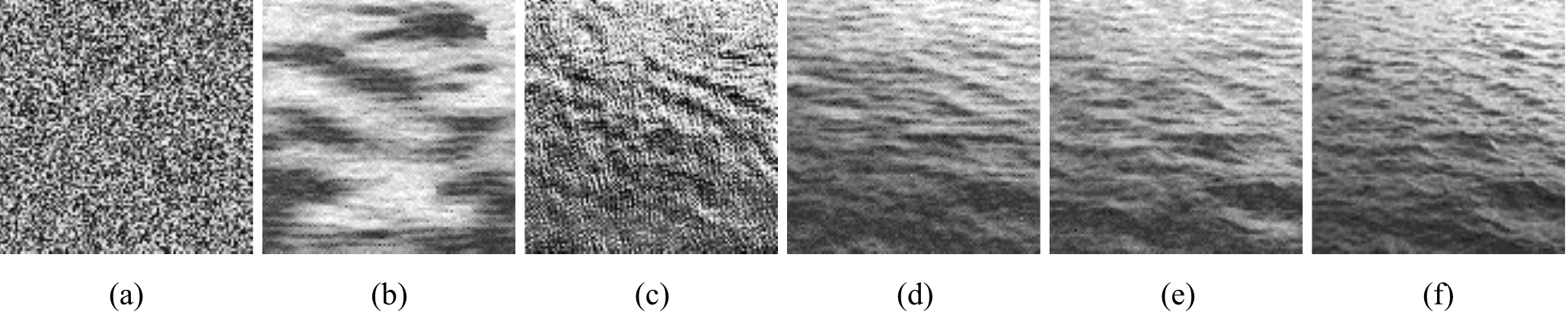}
\end{center}
   \caption{Synthesis for one frame of the ocean textured motion. (a) Initial
   uniform white noise image. (b) Synthesized frame with only static
   filters. (c) Synthesized frame with only motion filters. (d)
   Synthesized frame with both of static and motion filters. (e)
   Synthesized frame with all of the 3 types of filters. (f) The
   original observed frame.}
\label{evolution}
\end{figure*}

For implicit representation, the filters are $7\times7$ pixels in size and have $6$ scales, $12$ directions and $3$ speeds. Each type of filter has a special effect in textured motion synthesis, which will be discussed in section \ref{FRAME} and shown in Fig.\ref{evolution}.

\subsection{Learning motion primitives and reconstructing explicit regions}\label{learning}

\begin{figure}
\begin{center}
   \includegraphics[width=1\linewidth]{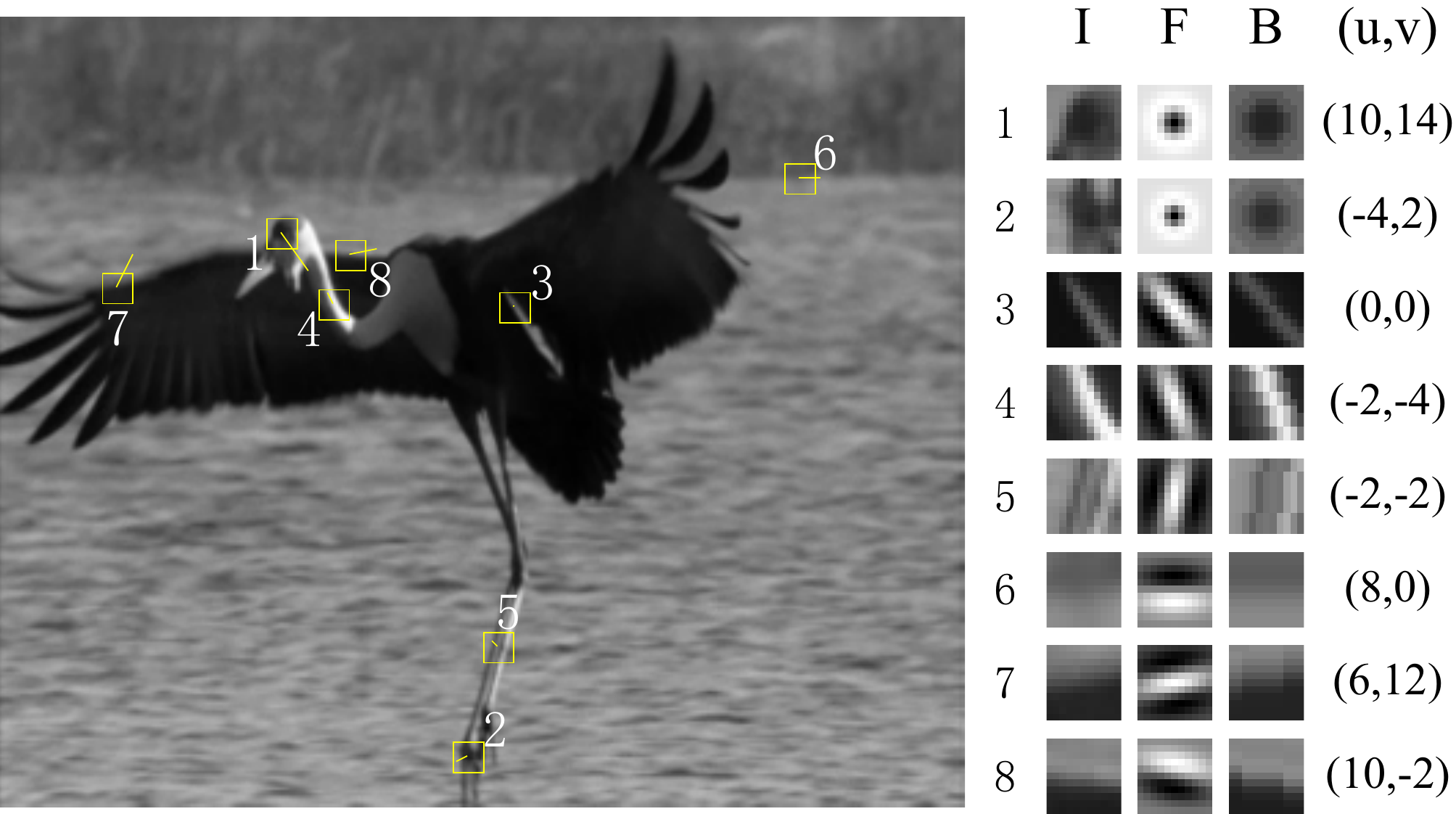}
\end{center}
   \caption{Some examples of primitives in a frame of video. Each group shows the original
   local image $\mathbf{I}$, the best fitted filter $\mathbf{F}$, the fitted primitive $\mathbf{B}\in\Delta_B$ and
   the velocity $(u,v)$, which represents the motion of $\mathbf{B}$.}
\label{tracking}
\end{figure}

\begin{figure*}[bp]
\begin{center}
   \includegraphics[width=0.9\linewidth]{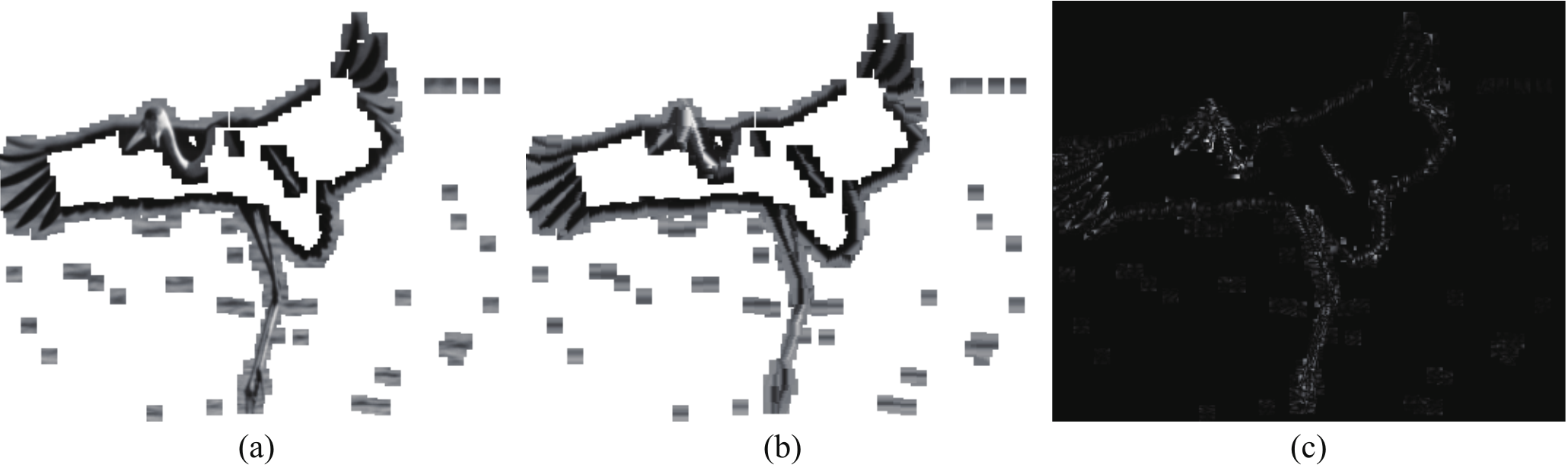}
\end{center}
   \caption{The reconstruction effect of sketchable regions by common primitives. (a) The observed frame. (b) The reconstructed frame. (c) The errors of reconstruction.}
\label{reconstruct}
\end{figure*}

After computing the sketchability and trackability maps of one frame, we extract explicit regions in the video. By calculating all the coefficients of each part with motion primitives from the primitive bank, $\alpha_{i,j}=\langle\mathbf{I}_{\Lambda_{tr,i}},B_j\rangle$, all the $\alpha_{i,j}$ are ranked from high to low. Each time, we select the primitive with the highest coefficient to represent the corresponding domain and then do local suppression to its neighborhood to avoid excessive overlapping of extracted domains. The algorithm is similar to matching pursuit (\cite{Mallat1993}) and the primitives are chosen one by one.

In our work, in order to alleviate computational complexity, $\alpha_{i,j}$ are calculated by filter responses. The filters used here are $11\times11$ pixels and have $18$ orientations and 8 scales. The fitted filter $F_j$ gives a raw sketch of the trackable patch and extracts property information, such as type and orientation, for generating the primitive. If the fitted filter is a Gabor-like filter, the primitive $B_j$ is calculated by averaging the intensities of the patch along the orientation of $F_j$, while if the fitted filter is a LoG-like filter, $B_j$ is calculated by averaging the intensities circularly around its center. Then $B_j$ is added to the primitive set $\mathbf{B}$ with its motion velocities calculated from the trackability map. It is also added into $\Delta_B$ for the dictionary buildup. The size of each primitive is $11\times11$, the same as the size of the fitted filter. And the velocity $(u,v)$ are two parameters for recording motion information. In Fig.\ref{primitive_example}, we show some examples of different types of primitives, such as blob, ridge and edge. Fig.\ref{tracking} shows some examples of reconstruction by motion primitives. In each group, the original local image, the fitted filter, the generated primitive and the motion velocity are given. In the frame, each patch is marked by a square with a short line for representing its motion information.

Through the matching pursuit process, the sketchable regions are reconstructed by a set of common primitives. Fig.\ref{reconstruct} shows an example of the sketchable region reconstruction by using a series of common primitives. By comparing the observed frame (a) and reconstructed frame (b), (c) shows the error of reconstruction. The more detailed quantitative assessment is given in section \ref{vps_syn}. It is evident that a rich dictionary of video primitives can lead to a satisfactory reconstruction of explicit regions of videos.

For non-sketchable but trackable regions, based on the trackability map, we get the motion trace of each local trackable patch. Because each patch cannot be represented by a shared primitive, we record the whole patch and motion information as a special primitive for video reconstruction. It is obvious that special primitives increase model complexity compared with common primitives. However, as stated in section \ref{ExRep}, the percentage of special primitives for the explicit region reconstruction of one video is very small (around 2-3\%), hence it will not affect the final storage space significantly.

%

\subsection{Synthesizing textured motions by ST-FRAME}\label{FRAME}

Each local volume $\mathbf{I}_{\Lambda_0}$ of textured motion located at $\Lambda_0$ follows a Markov random field model conditioned on its local neighborhood $\mathbf{I}_{\partial\Lambda_0}$ following (\ref{descriptivemodel}),
\begin{gather}
p(\mathbf{I}_{\Lambda_0}|\mathbf{I}_{\partial\Lambda_0};
\mathbf{F},\beta)\propto\exp\{-\sum_{k}{\langle\beta_k,H_k(\mathbf{I}_{\Lambda_0})\rangle}\},
\end{gather}
where Lagrange parameters $\beta_k=\{\beta_k^{(i)}\}_{i=1}^L\in\beta$ are the discrete form of potential function $\beta_k()$ learned from input videos by maximum likelihood,
\begin{gather}\label{mle}
\begin{aligned}
&\hat{\beta}=\arg\min\limits_{\beta}\log
p(\mathbf{I}_{\Lambda_0}|\mathbf{I}_{\partial\Lambda_0}; \beta,
\mathbf{F})\\&\hspace{0.5em}=\arg\max\limits_{\beta}\{-\log
Z(\beta)-\sum_{k}{<\beta_k,H_k(\mathbf{I}_{\Lambda_0})>}\}
\end{aligned}
\end{gather}
But the closed form of $\beta$ is not available in general. So it can be solved iteratively by
\begin{gather}\label{potential}
\frac{d\beta^{(i)}}{dt}=E_{p(\mathbf{I}; \beta,
\mathbf{F})}[H^{(i)}]-H^{obs(i)}
\end{gather}

In order to draw a typical sample frame from $p(\mathbf{I}; \mathbf{F},\beta)$, we use the Gibbs sampler which simulates a Markov chain. Starting from any random image, e.g. a white noise, it converges to a stationary process with distribution $p(\mathbf{I}; \mathbf{F},\beta)$. Therefore, we get the final converged results dominated by $p(\mathbf{I}; \beta, \mathbf{F})$, which characterizes the observed dynamic texture.

In summary, the process of textured motion synthesis is given by the
following algorithm.

\begin{flushleft}

\hrule

\hspace{0.1cm}

\textbf{Algorithm 1. Synthesis for Textured Motion by ST-FRAME}

\setlength{\hangindent}{1em} Input video
$\mathbf{I}^{obs}=\{\mathbf{I}_{(1)},...,\mathbf{I}_{(m)}\}$.

\setlength{\hangindent}{1em} Suppose we have
$\mathbf{I}^{syn}=\{\mathbf{I}_{(1)}^{syn},...,\mathbf{I}_{(m-1)}^{syn}\}$,
our goal is to synthesize the next frame $\mathbf{I}_{(m)}^{syn}$.

\setlength{\hangindent}{1em} Select a group of spatio-temporal
filters from a filter bank $\mathbf{F}=\{F_k\}_{k=1}^K\in\Delta_F$.

\setlength{\hangindent}{1em} Compute ${h_k, k=1,...,K}$ of
$\mathbf{I}^{obs}$.

\setlength{\hangindent}{1em} Initialize $\beta_k^{(i)}\leftarrow 0,
k=1,...,K, i=1,...,L$.

\setlength{\hangindent}{1em} Initialize $\mathbf{I}_{(m)}^{syn}$ as
a uniform white noise image.

Repeat

\hspace{0.5em}\setlength{\hangindent}{1.5em} Calculate $h_k^{syn},
k=1,2,...,K$ from $\mathbf{I}^{syn}$.

\hspace{0.5em}\setlength{\hangindent}{1.5em} Update $\beta_k,
k=1,...,K$ and $p(\mathbf{I}; \mathbf{F},\beta)$.

\hspace{0.5em}\setlength{\hangindent}{1.5em} Sample
$\mathbf{I}_{(m)}^{syn}\sim p(\mathbf{I}; \mathbf{F},\beta)$ by
Gibbs sampler.

\setlength{\hangindent}{1em} Until
$\frac{1}{2}\sum_{i=1}^L|h_k^{(i)}-h_k^{syn(i)}|\leq\epsilon$ for
$k=1,2,...,K$.

\hspace{0.1cm}

\hrule

\end{flushleft}

Fig.\ref{evolution} shows an example of the synthesis process. (f) is one frame from textured motion of ocean. Starting from a white noise frame in (a), (b) is synthesized with only 7 static filters. It shows high smoothness in spatial domain, but lacks temporal continuity with previous frames. However, in (c) the synthesis with only 9 motion filters has similar macroscopic distribution to the observed frame, but appears quite grainy over local spatial relationship. By using both static and motion filters, the synthesis in (d) performs well on both spatial and temporal relationships. Compared with (d), the synthesis by 2 extra flicker filters in (e), shows more smoothness and more similar to the observed frame.

\begin{figure}
\begin{center}
   \includegraphics[width=1\linewidth]{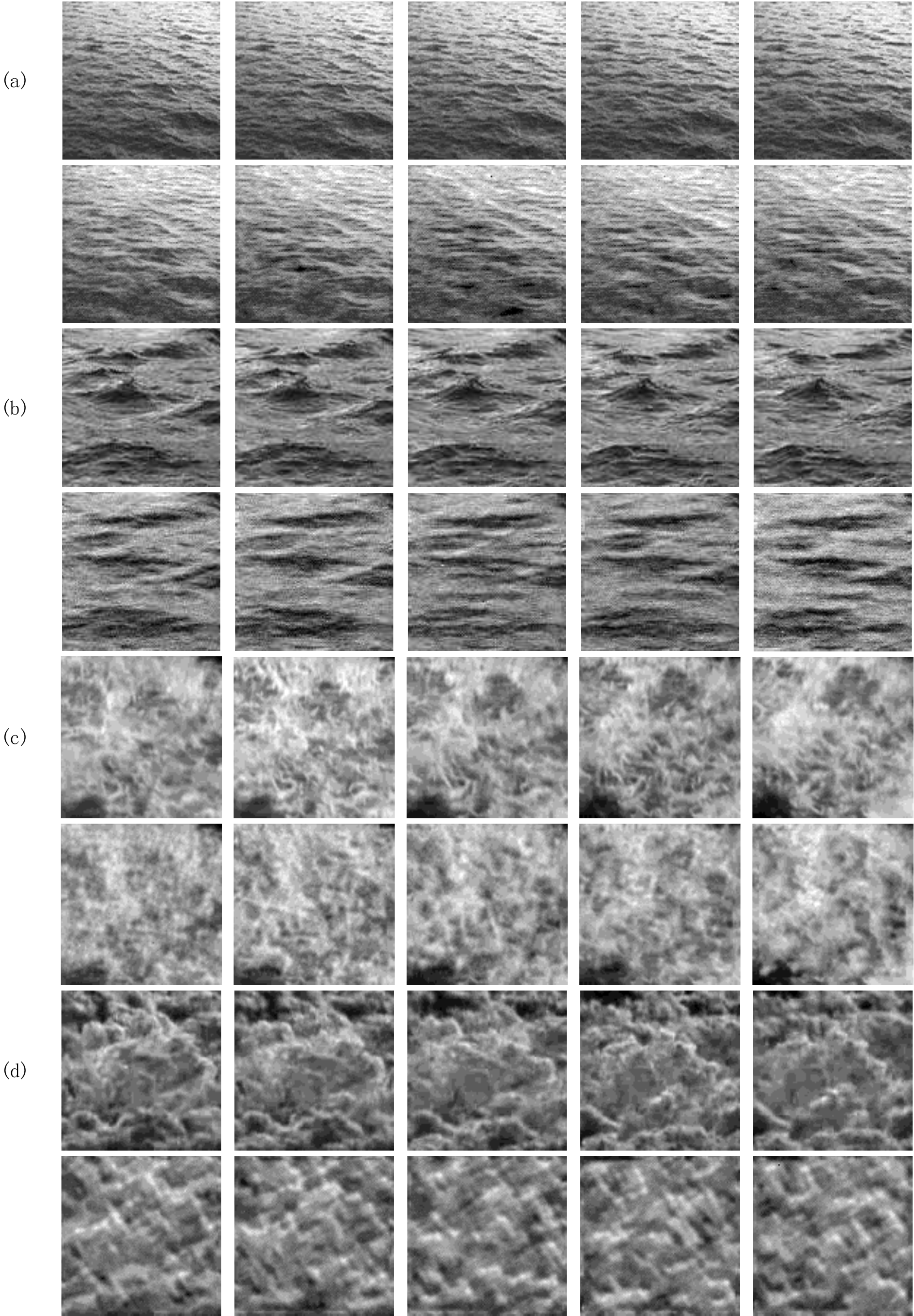}
\end{center}
   \caption{Textured motion synthesis examples. For each group, the top row are the original videos and the bottom row shows the synthesized
   ones. (a) Ocean. (b) Water wave. (c) Fire. (d) Forest.}
\label{texturesyn}
\end{figure}

In Fig.\ref{texturesyn}, we show four groups of textured motion (4 bits) synthesis by Algorithm 1: ocean (a), water wave (b), fire (c) and forest (d). In each group, as time passes, the synthesized frames are getting more and more different from the observed one. It is caused by the stochasticity of textured motions. Although the synthesized and observed videos are quite different on pixel level, the two sequences are perceived extremely identical by human after matching the histograms of a small set of filter responses. This conclusion can be further supported by perceptual studies in section \ref{percep_std}. Fig.\ref{hist} shows that as $\mathbf{I}_{(m)}^{syn}$ changes from white noise (Fig.\ref{evolution}(a)) to the final synthesized result (Fig.\ref{evolution}(e)), the histograms of filter responses become matched with the observed ones.

Table \ref{cmprsr1} shows the comparison of compression ratios between ST-FRAME and the dynamic texture model (\cite{Doretto2003}). It has a significantly better compression ratio than the dynamic texture model, because the dynamic texture model has to record PCA components as large as the image size.

\begin{figure}
\begin{center}
   \includegraphics[width=1\linewidth]{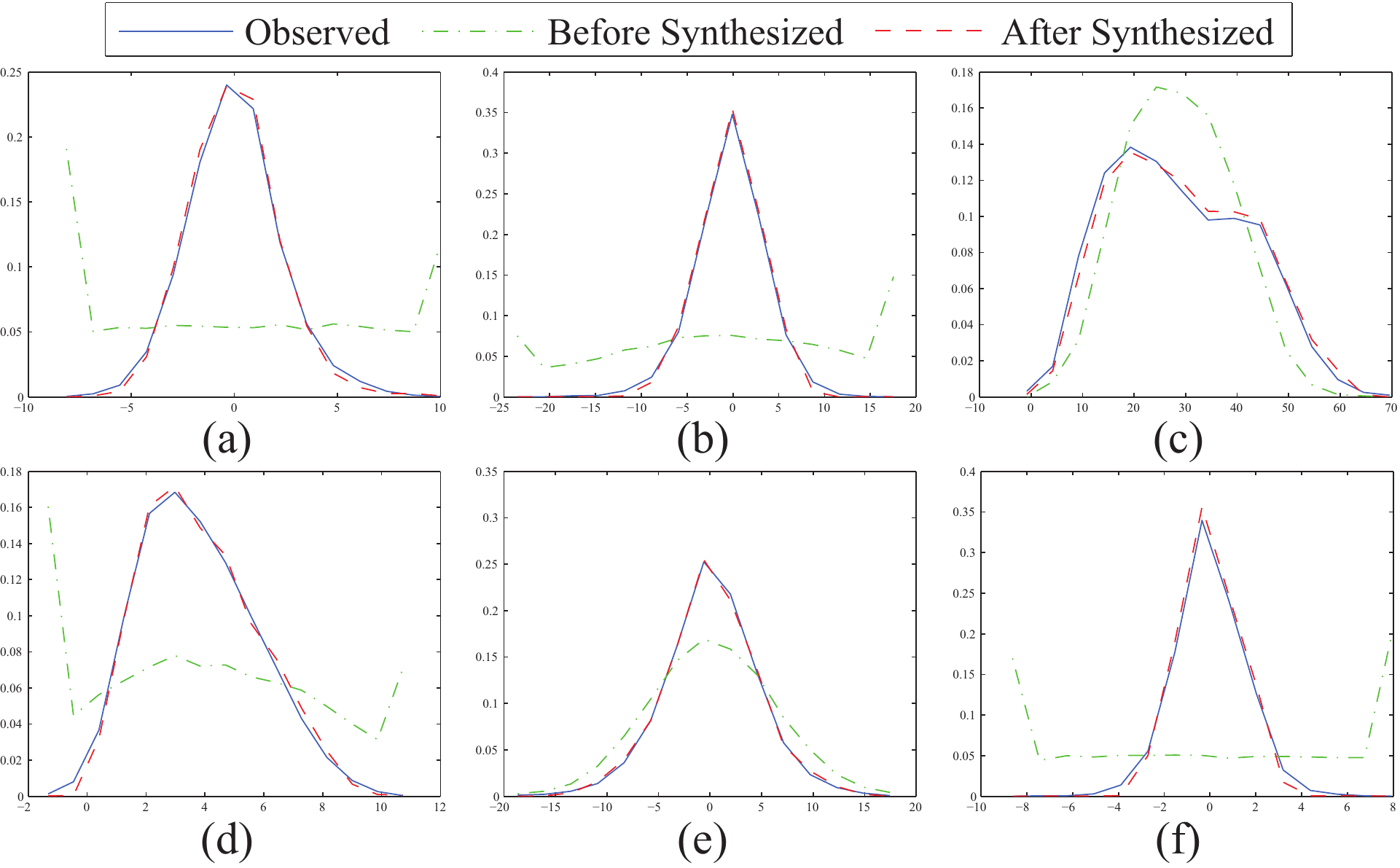}
\end{center}
   \caption{Matching of histograms of spatio-temporal filter responses for Ocean. The filters are
   (a) Static LoG(5$\times$5). (b) Static Gradient(vertical). (c) Motion Gabor(6,150$^\circ$).
   (d) Motion Gabor(2,60$^\circ$). (e) Motion Gabor(2,0$^\circ$). (f) Flicker LoG(5$\times$5).}\label{hist}
\end{figure}

\begin{table}
\caption{The number of parameters recorded and the compression ratios for synthesis of 5-frame textured motion videos by ST-FRAME and the dynamic texture model (\cite{Doretto2003}).}\label{cmprsr1}
\begin{center}
\begin{tabular}{|c|c|c|c|}
\hline
Example & Size & ST-FRAME & \tabincell{c}{Dynamic\\texture}\\
\hline
Ocean & $112\times112\times5$ & 558(0.89\%) & 25,096(40.01\%)\\
\hline
\tabincell{c}{Water\\wave} & $105\times105\times5$ & 465(0.84\%) & 22,058(40.01\%)\\
\hline
Fire & $110\times110\times5$ & 527(0.87\%) & 24,210(40.02\%)\\
\hline
Forest & $110\times110\times5$ & 465(0.77\%) & 24,210(40.02\%)\\
\hline
\end{tabular}
\end{center}
\end{table}

\subsection{Computing velocity statistics}

One popular method for velocity estimation is optical flow. Based on the optical flow, HOOF features extract the motion statistics by calculating the distribution of velocities in each region. Optical flow is an effective method for estimating the motions at trackable areas, but does not work for the intrackable dynamic texture areas. The three basic assumptions for optical flow equations, i.e.  brightness constancy between matched pixels in consecutive frames, smoothness among adjacent pixels and slow motion, are violated in these areas due to the stochastic nature of dynamic textures. Therefore, we go for a different velocity estimation method.

Considering one pixel $I(x,y,t)$ at $(x,y)$ in frame $t$, we denote its neighborhood as $I_{\partial\Lambda_{x,y,t}}$. Comparing patch $I_{\partial\Lambda_{x,y,t}}$ with all the patches in the previous frame within a searching radius, each patch corresponding to one velocity $v=(v_x,v_y)$, we obtain a distribution
\begin{gather}
p(v)\propto\exp\{-\Vert I_{\partial\Lambda_{x,y,t}}-I_{\partial\Lambda_{x-v_x,y-v_y,t-1}}\Vert^2\}
\end{gather}
This distribution describes the probability of the origin of the patch, i.e. the location where the patch $I_{\partial\Lambda_{x,y,t}}$ moves from. Equivalently, it reflects the average probability of the motions of the pixels in the patch. Therefore, by clustering all the pixels according to their velocity distribution, the cluster center of each cluster gives the velocity statistics of all the pixels in this cluster approximately, which reflects the motion pattern of these clustered pixels. Fig.\ref{v_global} and Fig.\ref{v_lcoal} show some examples of velocity statistics, in which the brighter, the higher probability while the darker, the lower probability. The meanings of these two figures are explained later.

Compared to HOOF, the estimated velocity distribution is more suitable for modeling textured motion. Firstly, the velocity distribution is estimated pixel-wisely. Hence it can depict more non-smooth motions. Secondly, although it seeks to compare the intensity pattern around a point to nearby regions at a subsequent temporal instance, which seems to also take brightness constancy assumption into account, the difference here is that it calculates the probability of motions rather than the single pixel correspondence. As a result, the constraints by the assumption is weakened, and it has the ability to represent stochastic dynamics.

\subsection{Synthesizing textured motions by MA-FRAME}

In MA-FRAME model, similar to ST-FRAME, each local volume $I_{\Lambda_0}$ of textured motion follows a Markov random field model. However, the  difference is that MA-FRAME extracts motion information via the distribution of velocities $v$.

In the sampling process, $I_{\Lambda_{im,j}}$ and $v_{\Lambda_{im,j}}$ are sampled simultaneously following the joint distribution  in (\ref{descriptivemodel2}),
\begin{gather}
p(\mathbf{I}_{\Lambda_{im,j}};\mathbf{F},\beta)
\propto\exp\{-\langle\mathbf{\beta},\mathbf{H}(\mathbf{I}_{\Lambda_{im,j}},\mathbf{v}_{\Lambda_{im,j}})\rangle\}.\notag
\end{gather}

In experiments, we design an effective way for sampling from the above model. For each pixel, we build a 2D-distribution matrix, whose two dimensions are velocities and intensities respectively, to guide the sampling process. The sampling probability for every candidate (labeled by one velocity and one intensity) is obtained by integrating motion score, appearance score and multiplying smoothness weight,
\begin{align}
SCORE(v,I) \notag\\
\propto\exp\{-\omega(v)&(\langle \beta^{(t)},\Vert H^{(t)}-H^{(t)obs}\Vert^2\rangle \notag\\
& +\langle \beta^{(s)},\Vert H^{(s)}-H^{(s)obs}\Vert^2\rangle)\}.\notag
\end{align}
\begin{figure}
\begin{center}
   \includegraphics[width=0.9\linewidth]{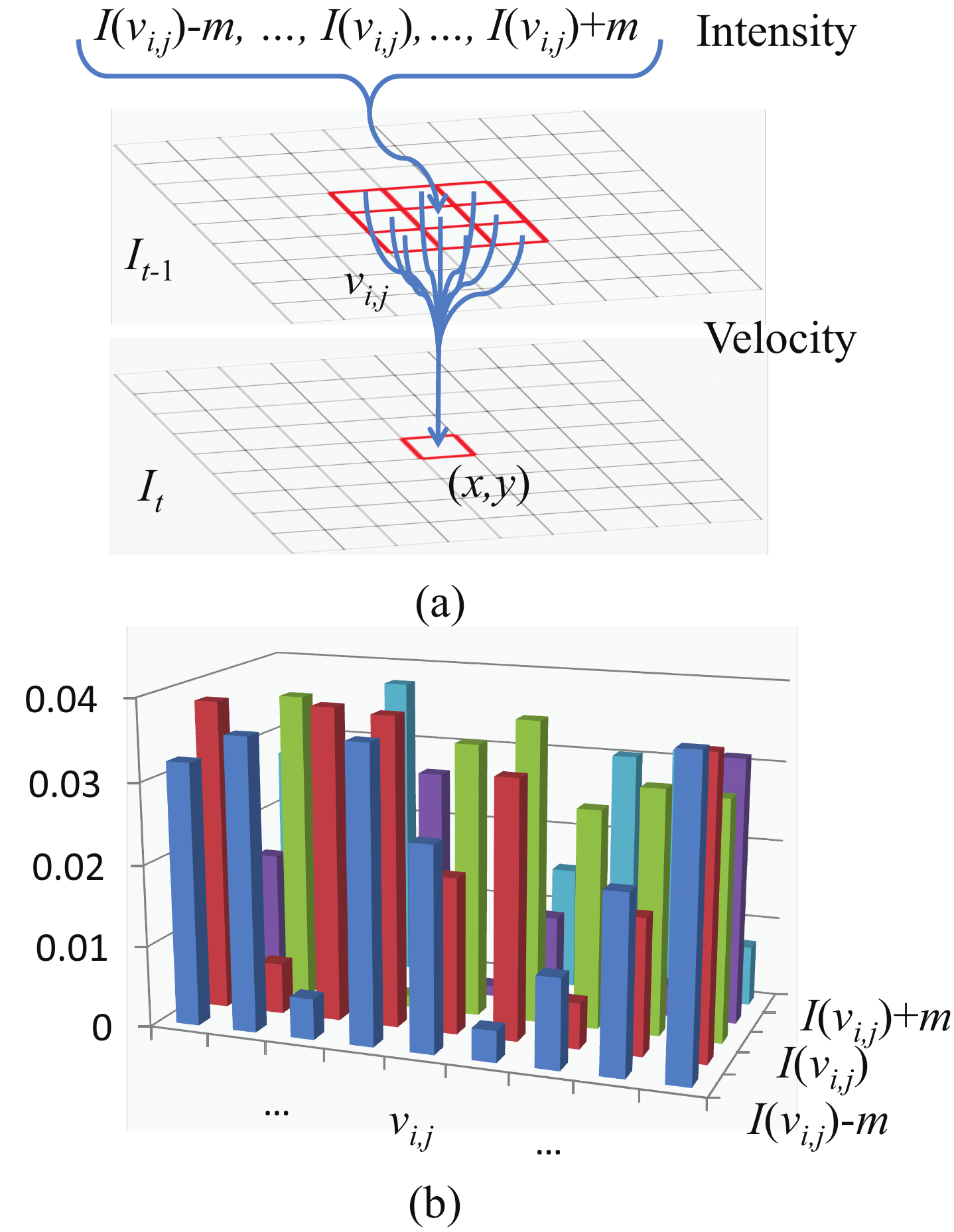}
\end{center}
   \caption{Sampling process of MA-FRAME. (a) For each pixel of current frame $I_t(x,y)$, the sample candidates are perturbation intensities of its neighborhood pixels in previous frame $I_{t-1}$ dominated by different velocities. (b) The velocity list and intensity perturbations construct two dimensions of the 2D distribution matrix, which is used for sampling $I_t(x,y)$. Here, $I(v_{i,j})$ is short for $I_{t-1}(x-v_{x,i},y-v_{y,j})$ and $i,j$ are indexes for different velocities in the neighborhood area.}\label{sample_v}
\end{figure}
\begin{figure*}
\begin{center}
   \includegraphics[width=1\linewidth]{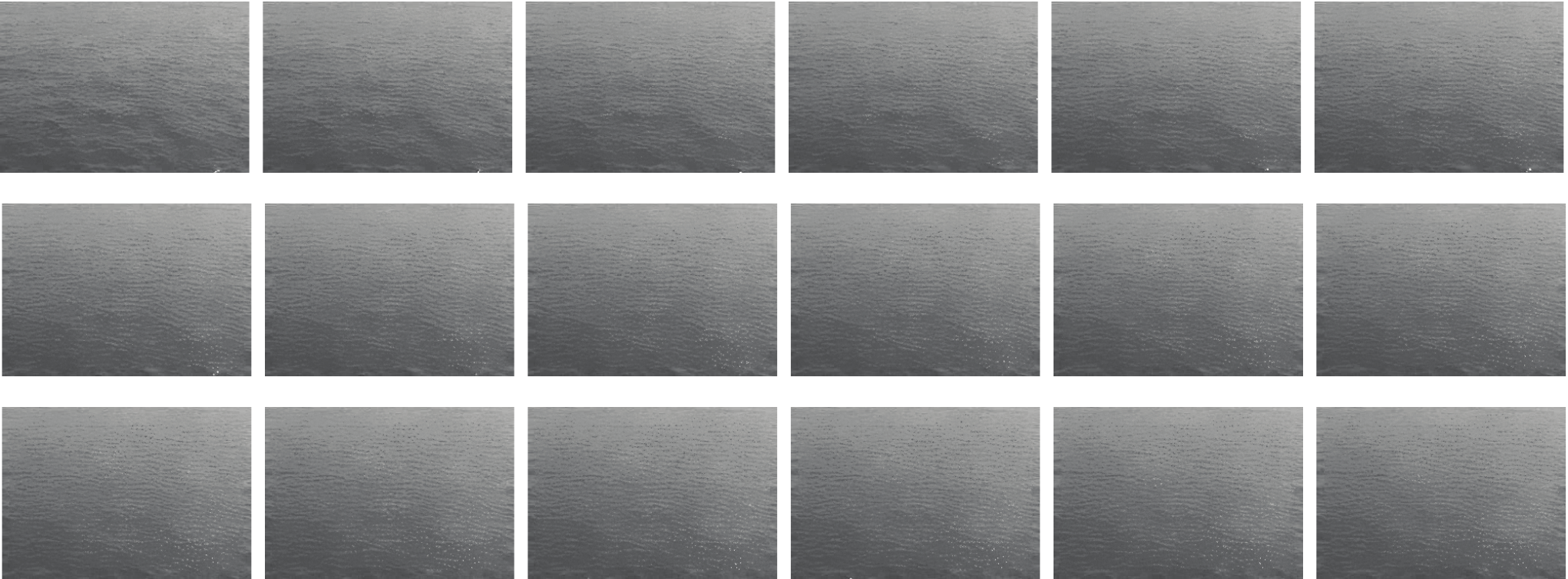}
\end{center}
   \caption{Texture synthesis for 18 frames of ocean video (from top-left to bottom-right) by MA-FRAME.}\label{texture_2_1}
\end{figure*}
\begin{figure*}
\begin{center}
   \includegraphics[width=1\linewidth]{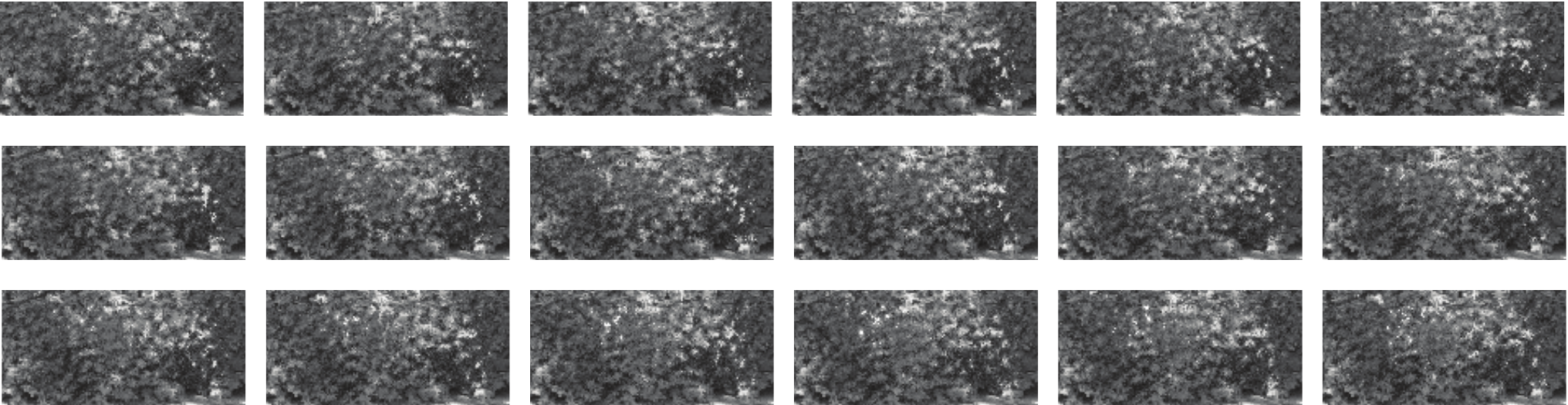}
\end{center}
   \caption{Texture synthesis for 18 frames of bushes video (from top-left to bottom-right) by MA-FRAME.}\label{texture_2_2}
\end{figure*}
The details are explained with the illustration by Fig.\ref{sample_v} for the sampling method at one pixel. For each pixel $(x,y)$ of the current frame $I_t$, we consider its every possible velocity $v_{i,j}=(v_{x,i},v_{y,j})$ within the range $-v_{max}\leq v_{x,i},v_{y,j}\leq v_{max}$. Each velocity corresponds to a position $(x-v_{x,i},y-v_{y,j})$ in the previous frame $I_{t-1}$. Under velocity $v_{i,j}$, the perturbation range of $I_{t-1}(x-v_{x,i},y-v_{y,j})$ yields the intensity candidates for $I_t(x,y)$ which is a  smaller interval than $[0,255]$ and thus saves computational complexity. In the shown example (Fig.\ref{sample_v}(a)), $v_{max}=1$ and the perturbed intensity range is $[I_{t-1}(x-v_{x,i},y-v_{y,j})-m, I_{t-1}(x-v_{x,i},y-v_{y,j})+m]$. Therefore we have $9$ velocity candidates and $2m+1$ intensity candidates for each velocity, hence the size of the sampling matrix is $9\times(2m+1)$ (Fig.\ref{sample_v}(b)).  With the motion constraints given by matching the velocity statistics, the velocity candidates have their motion scores. With the appearance constraints given by matching the filter response histograms, intensity candidates have their appearance scores. By integrating the two sets of scores, we obtain a preliminary sampling matrix shown in Fig.\ref{sample_v}(b).

In order to guarantee the motion of each pixel is as consistent as possible with its neighborhoods to make the macroscopic motion smooth enough, we add a set of weights on the distribution matrix, in which each multiplier for candidates of one velocity is calculated by
\begin{gather}
\begin{aligned}
&\omega(v_{x,i},v_{y,j})\notag\\
&=\sum_{(x_k,y_k)\in\partial\Lambda_{(x,y)}}\Vert v_{x,i}-v_{x_k}\Vert^2+\Vert v_{y,j}-v_{y_k}\Vert^2.\notag
\end{aligned}
\end{gather}
The weights encourage the velocity candidate which is closer to the velocities of its neighbours. With the weights, the sampled velocities are prone to be regarded as ``blurred'' optical flow. The main difference is that it preserves the uncertainty of dynamics in a texture motion, but not definite velocities of every local pixel.

After multiplying the weights to the preliminary matrix, we get the final sampling matrix. Although the main purpose of MA-FRAME is sampling intensities of each pixel from a textured motion, the sampling for intensities is highly related to velocities, and the sampling process is actually based on the joint distribution of velocity and intensity.

In summary, textured motion synthesis by MA-FRAME is given as follows

\begin{flushleft}

\hrule

\hspace{0.1cm}

\textbf{Algorithm 2. Synthesis for Textured Motion by MA-FRAME}

\setlength{\hangindent}{1em} Input video
$\mathbf{I}^{obs}=\{\mathbf{I}_{(1)},...,\mathbf{I}_{(n)}\}$.

\setlength{\hangindent}{1em} Suppose we have
$\mathbf{I}^{syn}=\{\mathbf{I}_{(1)}^{syn},...,\mathbf{I}_{(m-1)}^{syn}\}$,
our goal is to synthesize the next frame $\mathbf{I}_{(m)}^{syn}$.

\setlength{\hangindent}{1em} Select a group of static and flicker filters from a filter bank $\mathbf{F}=\{F_k\}_{k=1}^{K_s}\in\Delta_F$, where $K_s$ is the number of selected filters.

\setlength{\hangindent}{1em} Compute $\{h_k^{(s)}, k=1,...,K_s\}$, $\{h_k^{(v)}, k=1,...,K_v\}$ of
$\mathbf{I}^{obs}$, where $K_v$ is the number of velocity clusters.

\setlength{\hangindent}{1em} Initialize $\beta_k^{(i)}\leftarrow 0, k=1,...,K, i=1,...,L$.

\setlength{\hangindent}{1em} Initialize velocity vector $V=(v(x,y))_{(x,y)\in \Lambda}$ uniformly, and initialize $\mathbf{I}_{(m)}^{syn}$ by choosing intensities based on $V$.

Repeat

\hspace{0.5em}\setlength{\hangindent}{1.5em} Calculate $\{h_k^{syn(s)}, k=1,2,...,K_s\}$ and $\{h_k^{syn(v)}, k=1,2,...,K_v\}$ from $\mathbf{I}^{syn}$.

\hspace{0.5em}\setlength{\hangindent}{1.5em} Update $\beta_k, k=1,...,K$ and $p(\mathbf{I}; \mathbf{F},\beta)$.

\hspace{0.5em}\setlength{\hangindent}{1.5em} Sample $(\mathbf{I}_{(m)}^{syn},\mathbf{V}_{(m)}^{syn})\sim p(\mathbf{I},\mathbf{V}; \mathbf{F},\beta)$ by Gibbs sampler.

\setlength{\hangindent}{1em} Until
$\frac{1}{2}\sum_{i=1}^L|h_k^{(i)}-h_k^{syn(i)}|\leq\epsilon$ for
$k=1,2,...,K_s+K_v$.

\hspace{0.1cm}

\hrule

\end{flushleft}

Fig.\ref{texture_2_1} and Fig.\ref{texture_2_2} show two examples of textured motion synthesis by MA-FRAME. Different from the synthesis results by ST-FRAME, it can deal with videos of larger size, higher intensity level (8 bits here compared to 4 bits in ST-FRAME experiments) and more frames because of its smaller sample space and higher temporal continuity. Furthermore, it generates better motion pattern representations.

\begin{figure}
\begin{center}
   \includegraphics[width=0.9\linewidth]{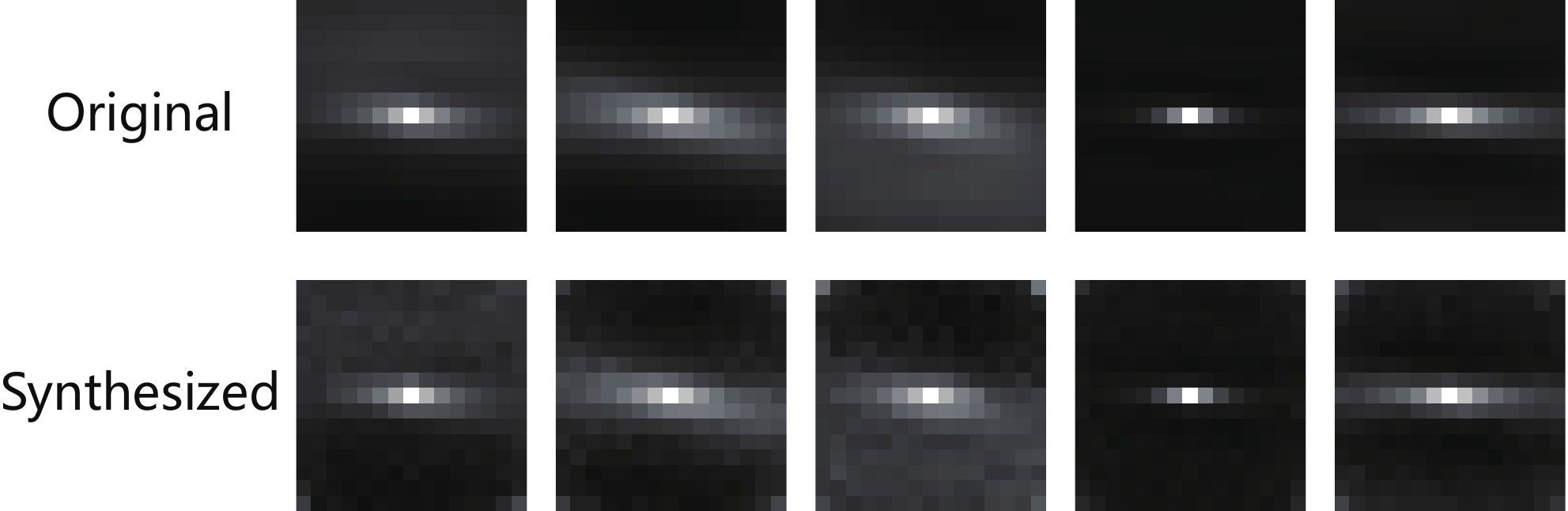}
\end{center}
   \caption{Five pairs of global statistics of velocities for comparison. Each patch corresponds to the neighborhood lattices as shown in Fig.\ref{sample_v}(a). The brighter means higher motion probability while the darker means lower probability.}\label{v_global}
\end{figure}

Fig.\ref{v_global} shows the comparison of velocities statistics between the original video and the synthesized video of different textured motion clusters, the brighter, the higher motion probability while the darker, the lower probability. It is easy to tell that they are quite consistent, which means the original and synthesized videos have similar macroscopical motion properties.

\begin{figure}
\begin{center}
   \includegraphics[width=0.9\linewidth]{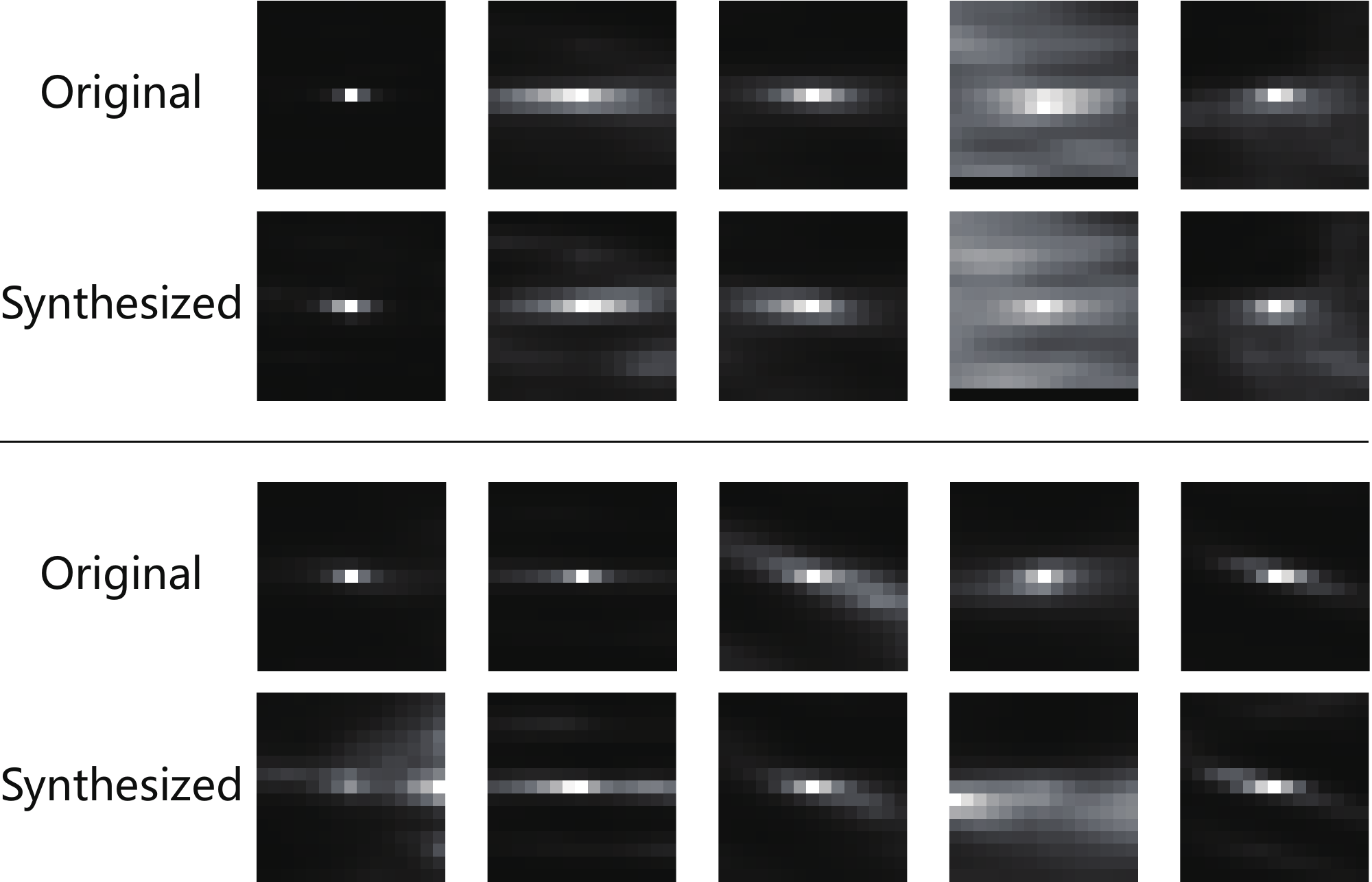}
\end{center}
   \caption{Ten pairs of local statistics of velocities for comparison. Upper row: original; lower row: synthesis. Each patch corresponds to the neighborhood lattices as shown in Fig.\ref{sample_v}(a). The brighter means higher motion probability while the darker means lower probability.}\label{v_lcoal}
\end{figure}

We also test local motion consistency between observed and synthesized videos by comparing velocity distributions of every pair of corresponding pixels. Fig.\ref{v_lcoal} shows the comparisons of ten pairs of randomly chosen pixels. Most of them match well. It demonstrated that the motion distributions of most of local patches also preserve well during the synthesis procedure.

\subsection{Dealing with occlusion parts in texture synthesis}

Before providing the full version of computational algorithm for VPS, we first introduce how to deal with occluded areas.

In video, dynamic background textures are often occluded by the movement of foreground objects. Synthesizing background texture by ST-FRAME uses histograms of spatio-temporal filter responses. When a textured region becomes occluded, the pattern no longer belongs to the same equivalence class. In this event, the spatio-temporal responses are not precise enough for matching the given histograms, and may cause a deviation in the synthesis results. These errors may accumulate over frames and the synthesis will ultimately degenerate completely. Synthesis by MA-FRAME has a greater problem because the intensities in the current frame are selected from small perturbations in intensities from the previous frame. If a pixel cannot be found from the neighborhood in the previous frame that belongs to the same texture class, the intensity it adopts may be incompatible with other pixels around it.

In order to solve this problem, occluded pixels are sampled separately by the original (spatial) FRAME model, which means, we have two classes of filter response histograms
\begin{itemize}
\item[1] Static filter response histograms $H^S$. Histograms are calculated by summarizing static filter responses of all the textural pixels;

\item[2] Spatio-temporal filter response histograms $H^{ST}$. Histograms are calculated by summarizing spatio-temporal filter responses of all the non-occluded textured pixels.
\end{itemize}

Therefore, in the sampling process, the occluded pixels and non-occluded pixels are treated differently. First, their statistics are constrained by different sets of filters; second, in MA-FRAME, the intensities of non-occlude pixels are sampled from the intensity perturbation of their neighborhood locations in previous frame, while the intensities of occluded pixels are sampled from the whole intensity space, say 0-255 for 8 bits grey levels.


\subsection{Synthesizing videos with VPS}\label{vps_syn}

In summary, the full version of the computational algorithm for video synthesis of VPS is presented as follows.

\begin{flushleft}

\hrule

\hspace{0.1cm}

\textbf{Algorithm 3. Video Synthesis via Video Primal Sketch}

\setlength{\hangindent}{1em} Input a video $\mathbf{I}^{obs}=\{\mathbf{I}_{(1)}^{obs},...,\mathbf{I}_{(m)}^{obs}\}$.

\setlength{\hangindent}{1em} Compute sketchability and trackability
for separating $\mathbf{I}^{obs}$ into explicit region
$\mathbf{I}_{\Lambda_{ex}}$ and implicit region
$\mathbf{I}_{\Lambda_{im}}$.

\setlength{\hangindent}{1em}\textbf{for} t=1:m

\hspace{1em}\setlength{\hangindent}{2em} Reconstruct $\mathbf{I}_{(t)\Lambda_{ex}}^{obs}$
by the sparse coding model with the selected primitives $\mathbf{B}$
chosen from the dictionary $\Delta_B$ to get $\mathbf{I}_{(t)\Lambda_{ex}}^{syn}$.

\hspace{1em}\setlength{\hangindent}{2em} For each region of homogeneous textured
motion $\Lambda_{im,j}$, using $\mathbf{I}_{(t)\Lambda_{ex}}^{syn}$ as boundary condition,
synthesize $\mathbf{I}_{(t)\Lambda_{im,j}}^{obs}$ by ST-FRAME model or MA-FRAME with the
selected filter set $\mathbf{F}$ chosen from the filter bank $\Delta_F$ to get $\mathbf{I}_{(t)\Lambda_{im}}^{syn}$.

\hspace{1em}\setlength{\hangindent}{2em} The synthesis of the $i$th frame of the video $\mathbf{I}_{(t)}^{syn}$ is given by aligning $\mathbf{I}_{(t)\Lambda_{ex}}^{syn}$ and $\mathbf{I}_{(t)\Lambda_{im}}^{syn}$ together seamlessly.

\setlength{\hangindent}{1em}\textbf{end for}

\setlength{\hangindent}{1em} Output the synthesized video $\mathbf{I}^{syn}$.

\hspace{0.1cm}

\hrule

\end{flushleft}

\begin{figure}
\begin{center}
   \includegraphics[width=1\linewidth]{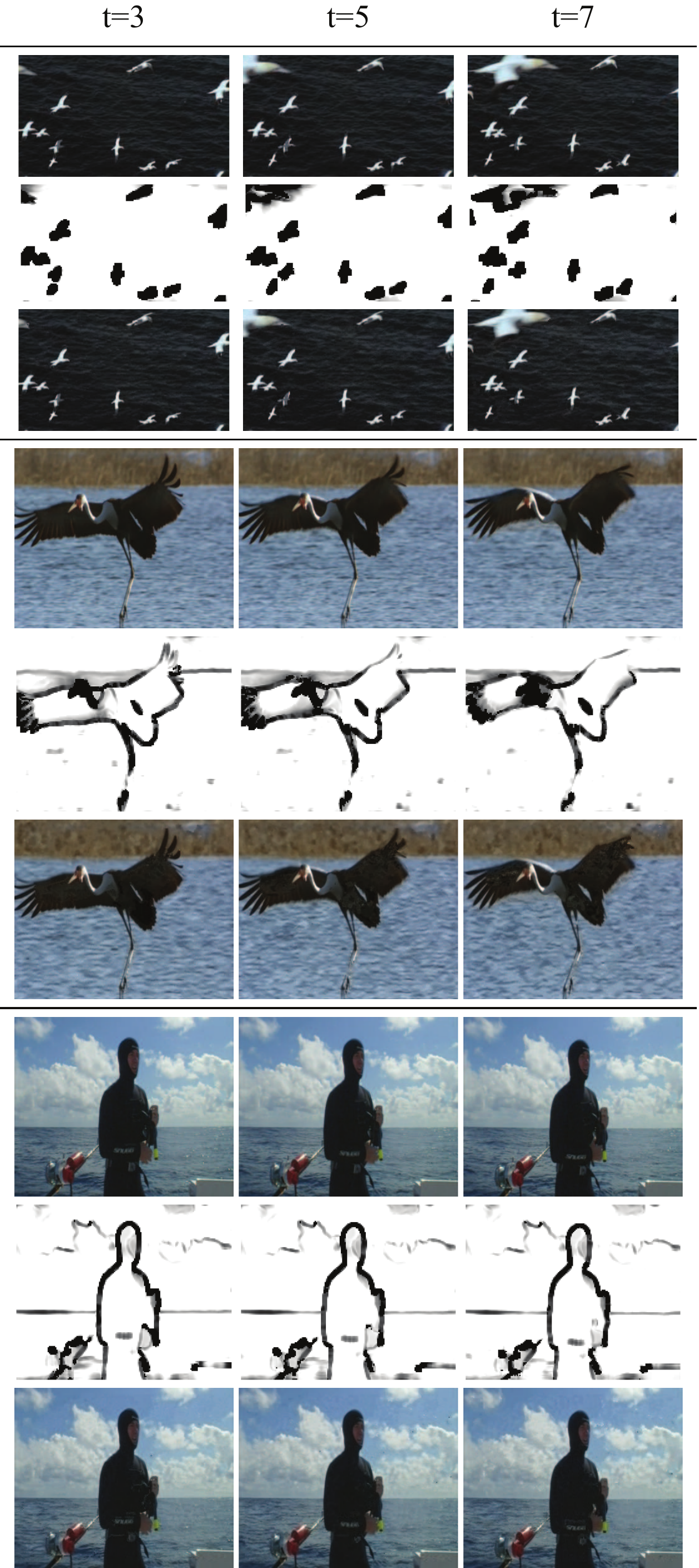}
\end{center}
   \caption{Video synthesis. For each experiment, Row 1: original frames; Row 2: trackability maps; Row 3: synthesized frames.}
\label{synthesis}
\end{figure}

Fig.\ref{illustration} shows this process as we introduced in section \ref{intro}. Fig.\ref{synthesis} shows three examples of video synthesis (YCbCr color space, 8 bits for grey level) by VPS frame by frame. In every experiment, observed frames, trackability maps, and final synthesized frames are shown. In Table \ref{cmprsr2}, H.264 is selected as the reference of compression ratio compared with VPS, from which we can tell VPS is competitive with state-of-art video encoder on video compression.

For assessing the quality of the synthesized results quantitatively, we adopt two criteria for different representations, rather than the traditional approach based on error-sensitivity as it has a number of limitations (\cite{ZWang2004}). The error for explicit representations is measured by the difference of pixel intensities
\begin{gather}
err^{ex}=\frac{1}{|\Lambda_{ex}|}\sum_{\Lambda_{ex}}\Vert I^{syn}-I^{obs}\Vert.
\end{gather}
while for implicit representations, the error is given by the difference of filter response histograms,
\begin{gather}
err^{im}=\frac{1}{n_{im}\times K}\sum_{n_{im},K}\Vert H_k(I_{\Lambda_{im,j}}^{syn})-H_k(I_{\Lambda_{im,j}}^{obs})\Vert.
\end{gather}
Table \ref{accuracy} shows the quality assessments of the synthesis, which demonstrates good performance of VPS on synthesizing videos.

\begin{table}[h]
\caption{Compression ratio of video synthesis by VPS and H.264 to
raw image sequence.}\label{cmprsr2}
\begin{center}
\begin{tabular}{|c|c|c|c|}
\hline
Example & Raw (Kb) & VPS (Kb) & H.264 (Kb) \\
\hline
1 & 924 & 16.02 (1.73\%) & 20.8 (2.2\%)\\
\hline
2 & 1,485 & 26.4 (1.78\%) & 24 (1.62\%)\\
\hline
3 & 1,485 & 28.49 (1.92\%) & 18 (1.21\%)\\
\hline
\end{tabular}
\end{center}
\end{table}

\begin{table}[h]
\caption{Error assessment of synthesized videos.}\label{accuracy}
\begin{center}
\begin{tabular}{|c|c|c|c|}
\hline
Example & Size(Pixels) & Error($I_{\Lambda_{ex}}$) & Error($I_{\Lambda_{im}}$)\\
\hline
1 & $190\times330\times7$ & 5.37\% & 0.59\%\\
\hline
2 & $288\times352\times7$ & 3.07\% & 0.16\%\\
\hline
3 & $288\times352\times7$ & 2.8\% & 0.17\%\\
\hline
\end{tabular}
\end{center}
\end{table}

\subsection{Computational Complexity Analysis}

In this subsection, we analyze the computational complexity of the algorithms studies in this paper. We discuss the complexity for four algorithms in the following. The implementation environment is the desktop computer with Intel Core i7 2.9 GHz CPU, 16GB memory and Windows 7 operating system.

\textbf{1) Video modeling by VPS.} Suppose one frame of a video contains $N$ pixels, of which, $N_{ex}$ pixels belong to explicit regions and $N_{im}$  in implicit regions. Let the size of the filter dictionary be $N_F$ and the filter size be $S_F$, the computational complexity for calculating filter responses is $O(NN_FS_F)$. For extracting and learning explicit bricks, the complexity is no more than $O(N_{ex}\sqrt{S_F})$. For calculating the response histograms of $K$ chosen filters within the implicit regions, the complexity is no more than $O(N_{im}Kk)$ if there are $k$ homogeneous textural area in the regions. To sum up, the total computation complexity for video coding is no more than $O(NN_FS_F+N_{ex}\sqrt{S_F}+N_{im}Kk)$. In our experiments, for coding one frame of the video with the size of $288\times352$, the time consumption is less than 0.5 seconds.

\textbf{2) Reconstruction of explicit regions.} Because the information of all the basis for explicit regions are recorded and there needs no additional computations for reconstructing, the computational complexity can be regarded as $O(1)$ and the reconstruction costs no time in comparison to other components.

\textbf{3) Synthesis of implicit regions by Gibbs sampling by ST-FRAME.} For one round sampling, each of the $N_{im}$ pixels will be sampled in the range of the overall intensity levels, say $L$. For every sampling candidate, i.e. one intensity, the score is calculated via the change of synthesized filter response histograms. To reduce the computation burden, we can simply update the change of filter responses caused by the change of the intensity on the current pixel. This operation requires $KS_F$ times of multiplications. As a result, the computational complexity for one round sampling of one frame is $O(N_{im}LKS_F)$. In the experiments of this paper, one frame will be sampled for about 20 rounds. Then the running time is about 2 minutes if the image is 4 bits and the size of implicit region is $10,000$ pixels.

\textbf{4) Synthesis of implicit regions by Gibbs sampling by MA-FRAME.} The computational complexity of MA-FRAME is quite similar with ST-FRAME. The biggest difference is the number of  sampling candidates. As the number of velocity candidates is $N_v$ and the intensity perturbation range is $[-m,m]$, the computational complexity is $O(N_{im}N_vmKS_F)$, which is on the same level with ST-FRAME. However, in real application, because the intensities of the neighborhood of one pixel are not far away, the intensities of the candidates with different velocities is quite redundant. As a result, MA-FRAME may save a lot of time compared with ST-FRAME, especially when the intensity level is high. For one frame with 8 bits and $60,000$ pixels, the running time is about 4 minutes within 20 rounds sampling.

In summary, the computational complexity of video modeling / coding by VPS is small, but that of video synthesis is quite large. It is because of texture synthesis procedure. In VPS, the textures are modeled by MRF and synthesized via a Gibbs sampling process, which is well known as a computational costing method. However, the video synthesis is only one of the applications of VPS and is used for verifying the correctness of the model. As a result, it is not the very important issue we care about here.

\subsection{Perceptual Study}\label{percep_std}

The error assessment of VPS is consistent with human perception. To support this claim, in this subsection, we present a series of human perception experiments and explore the relationship between perception accuracy. In the experiments below, the 30 participants include graduate students and researchers from mathematics, computer science and medical science. The age range is from 22 to 39, and they all have normal or corrected-to-normal vision.

\begin{figure}
\begin{center}
   \includegraphics[width=1\linewidth]{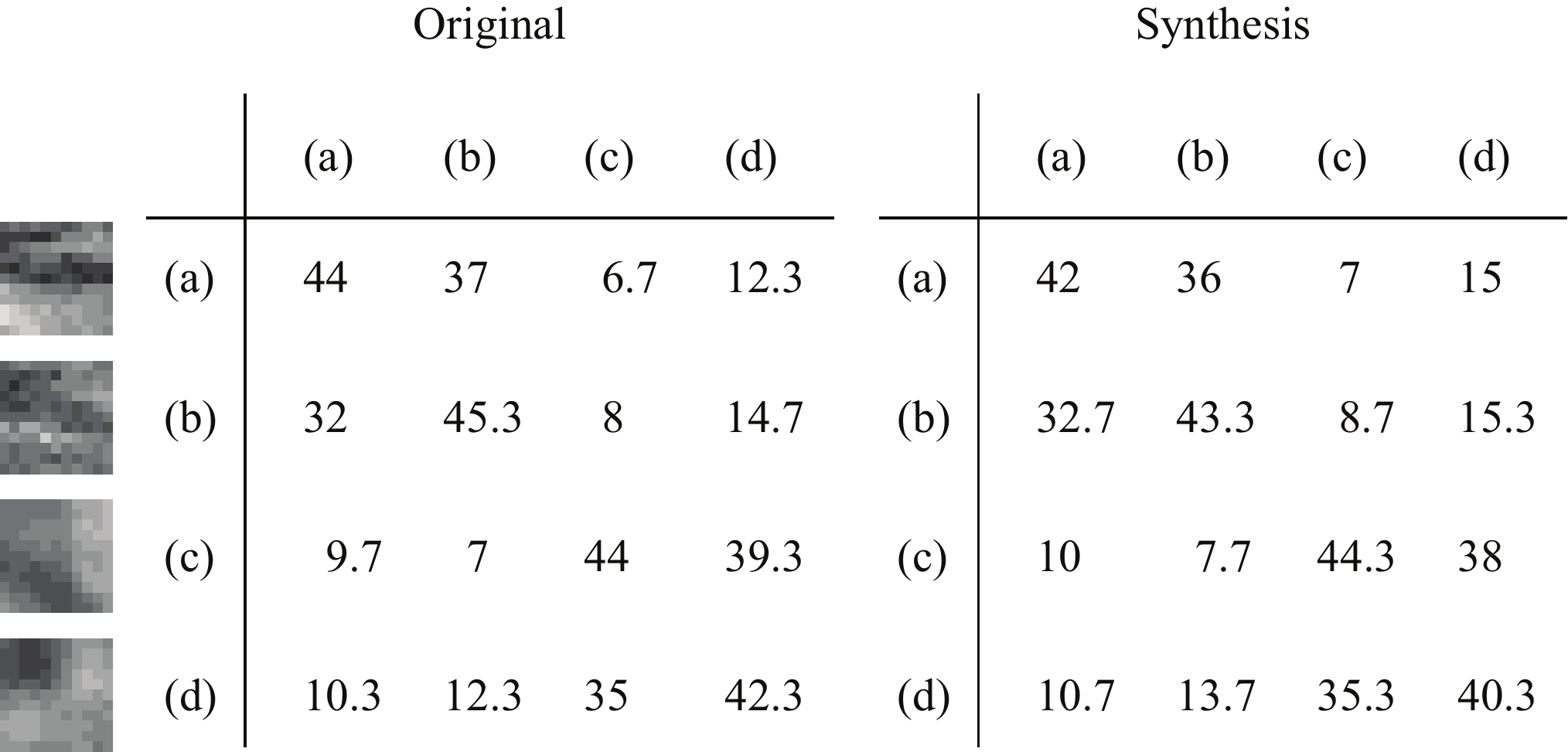}
\end{center}
   \caption{Perceptual confusion test on original and synthesized textured motions respectively. The size of cropped examples are $5\times5$.}
\label{phyche_5}
\end{figure}

\begin{figure}
\begin{center}
   \includegraphics[width=1\linewidth]{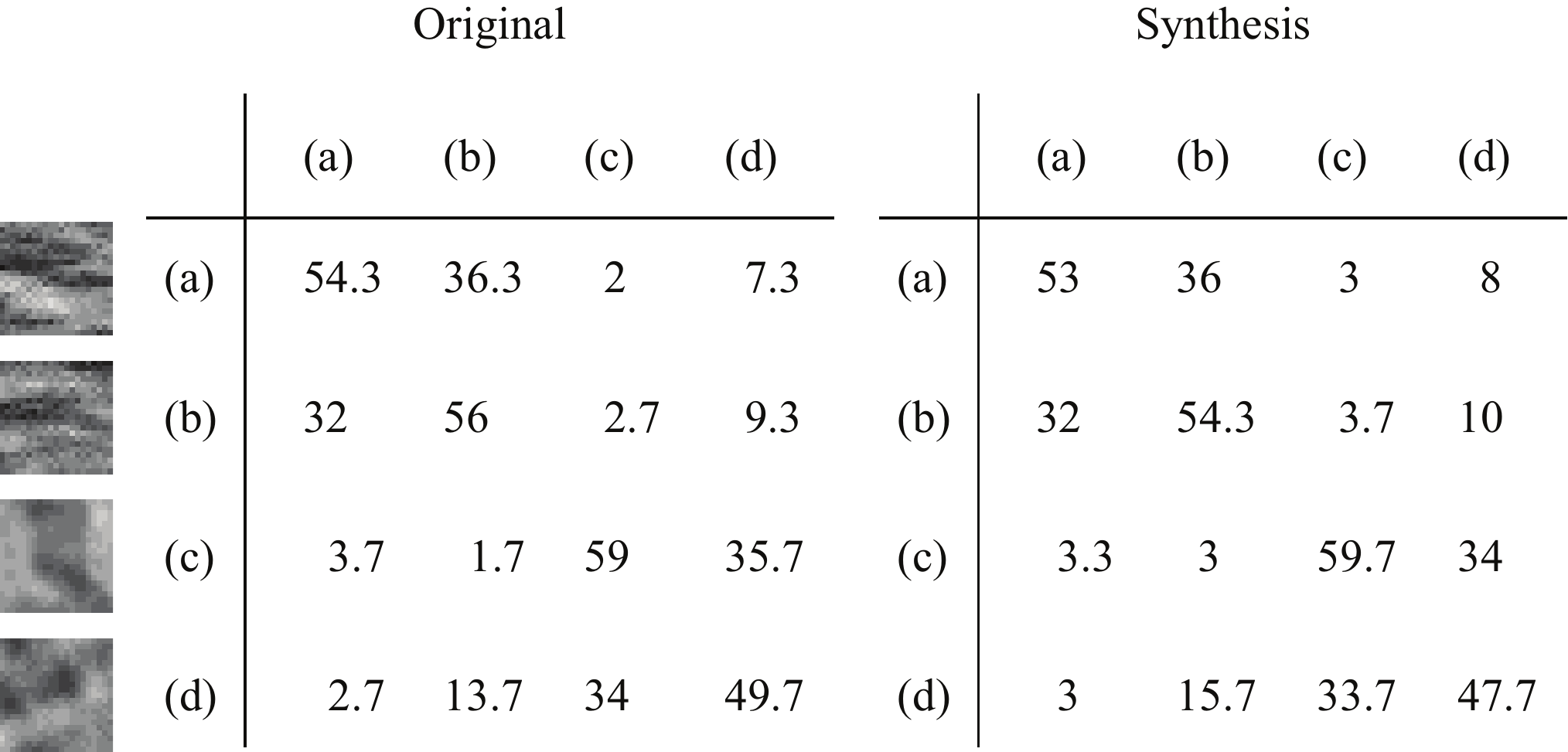}
\end{center}
   \caption{Perceptual confusion test on original and synthesized textured motions respectively. The size of cropped examples are $10\times10$.}
\label{phyche_10}
\end{figure}

\begin{figure}
\begin{center}
   \includegraphics[width=1\linewidth]{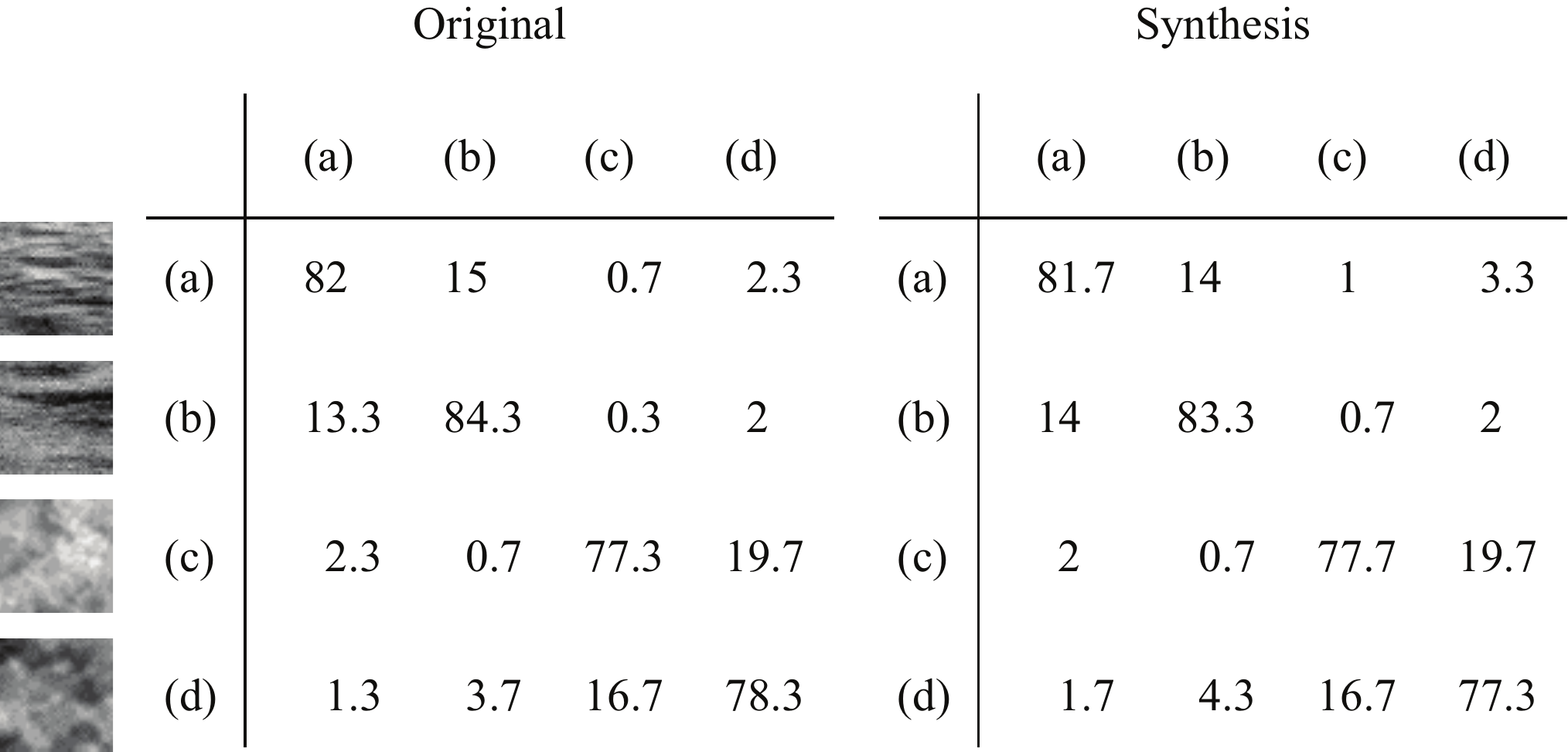}
\end{center}
   \caption{Perceptual confusion test on original and synthesized textured motions respectively. The size of cropped examples are $20\times20$.}
\label{phyche_20}
\end{figure}

In the first experiment, we randomly crop several clips of videos with different sizes from the four synthesized textured motion examples and their corresponding original videos (as shown in the left side of Fig. \ref{phyche_5}, \ref{phyche_10} and \ref{phyche_20}, each video is shown one frame as an example which is marked by (a), (b), (c) and (d) respectively, and they are in different sizes but shown in the same size after zooming for better shows). And then for original and synthesized examples respectively, each participant is shown 40 clips one by one (10 clips from each texture) and is required to guess which texture they come from. We show 3 representative groups of results below for demonstration, in which the sizes of cropped examples are $5\times5$, $10\times10$ and $20\times20$ respectively. Both of the confusion rates (\%) of original and synthesized examples are shown in the tables on the right side in Fig. \ref{phyche_5}, \ref{phyche_10} and \ref{phyche_20}. Each row gives the average confusion rates, which the video clip labeled by the row title is judged coming from textures labeled by the column titles. In order to test if the syntheses are perceived the same with the original videos, we compare the original and synthesis confusion tables in each group. From the results, we can tell that the confusion tables are mostly consistent. For more precise quantitative estimation, we also analyze the recognition accuracies by ANOVA in Table \ref{phyche_anova}, in which, each row shows the corresponding $F$ and $p$ values for each texture in all the three groups. The results show that the recognition accuracies on original and synthesized textures do not differ significantly.

\begin{table}[h]
\caption{The ANOVA results of analyzing recognition accuracies of original and synthesized textures. For each texture in every group, the corresponding $F$ and $p$ values are shown respectively.}\label{phyche_anova}
\begin{center}
\begin{tabular}{|c|c|c|c|}
\hline
F/p & Group 1 & Group 2 & Group 3\\
\hline
(a) & 1.34/0.2520 & 0.65/0.4222 & 0.02/0.8813\\
\hline
(b) & 0.96/0.3305 & 0.70/0.4065 & 0.20/0.6583\\
\hline
(c) & 0.06/0.8100 & 0.15/0.6993 & 0.03/0.8563\\
\hline
(d) & 1.43/0.2366 & 1.08/0.3088 & 0.26/0.6151\\
\hline
\end{tabular}
\end{center}
\end{table}

Also, it is noted that texture (a) and (b) appear similarly while (c) and (d) tend to be confused with each other. Therefore, the confusion rates between (a) and (b), (c) and (d) are apparently larger. However, from Fig. \ref{phyche_5} to \ref{phyche_20}, as the size of cropped videos gets larger, the confusion rate becomes lower, and actually when the size goes larger than $25\times25$ in this experiment, the accuracies get very close to 100\%. This experiment demonstrates the fact that the dynamic textures synthesized by the statistics of dynamic filters can be well discriminated by human vision, although the synthesized one and the original one are totally different on pixel level. Therefore it is evident that the approximation of filter response histograms reflects the quality of video synthesis. Furthermore, it is proved that larger area textures give much better perception effect because human can extract more macroscopic statistical information and motion-appearance characteristics, while small size local areas can only provide salient structural information which may be shared by a various of different videos.

\begin{table}[h]
\caption{The accuracy of differentiating the original video from the synthesized one in different scales. As the percentage is getting closer to 50\%, it means it is harder to discriminate the original and synthesized videos for observers.}\label{phyche_video}
\begin{center}
\begin{tabular}{|c|c|c|c|c|}
\hline
Video & Scale 100\% & Scale 75\% & Scale 50\% & Scale 25\%\\
\hline
1 & 66.7 & 56.7 & 46.7 & 50\\
\hline
2 & 100 & 90 & 73.3 & 63.3\\
\hline
3 & 73.3 & 63.3 & 50 & 53.3\\
\hline
\end{tabular}
\end{center}
\end{table}

In the second experiment, we test if the synthesized video by VPS gives similar vision impact compared with the original video. Each time we provide the original and the synthesized videos to one participant in the same scale. The videos are played synchronously and the participants are required to point out which is the original video in 5 seconds. Each pair of videos is tested in four scales, 100\%, 75\%, 50\% and 25\%. The accuracy are shown in Table \ref{phyche_video}. From the result, when the videos are shown in larger scales, it is easier to discriminate the original and synthesized videos, because a lot of structural details can be noticed by the observers. But as the scale gets smaller, the macroscopic information gives the major impact to the vision system, therefore the original and synthesized video are perceived almost the same, so that the accuracy get lower and approach to 50\%. From this experiment, it is evident that although VPS cannot give the complete reconstruction of a video on pixel level, especially for dynamic textures, but the synthesis gives human similar vision impact, which means most of the key information for perception are kept via VPS model.

\subsection{VPS adapting over scales, densities and dynamics}

\begin{figure}
\begin{center}
   \includegraphics[width=1\linewidth]{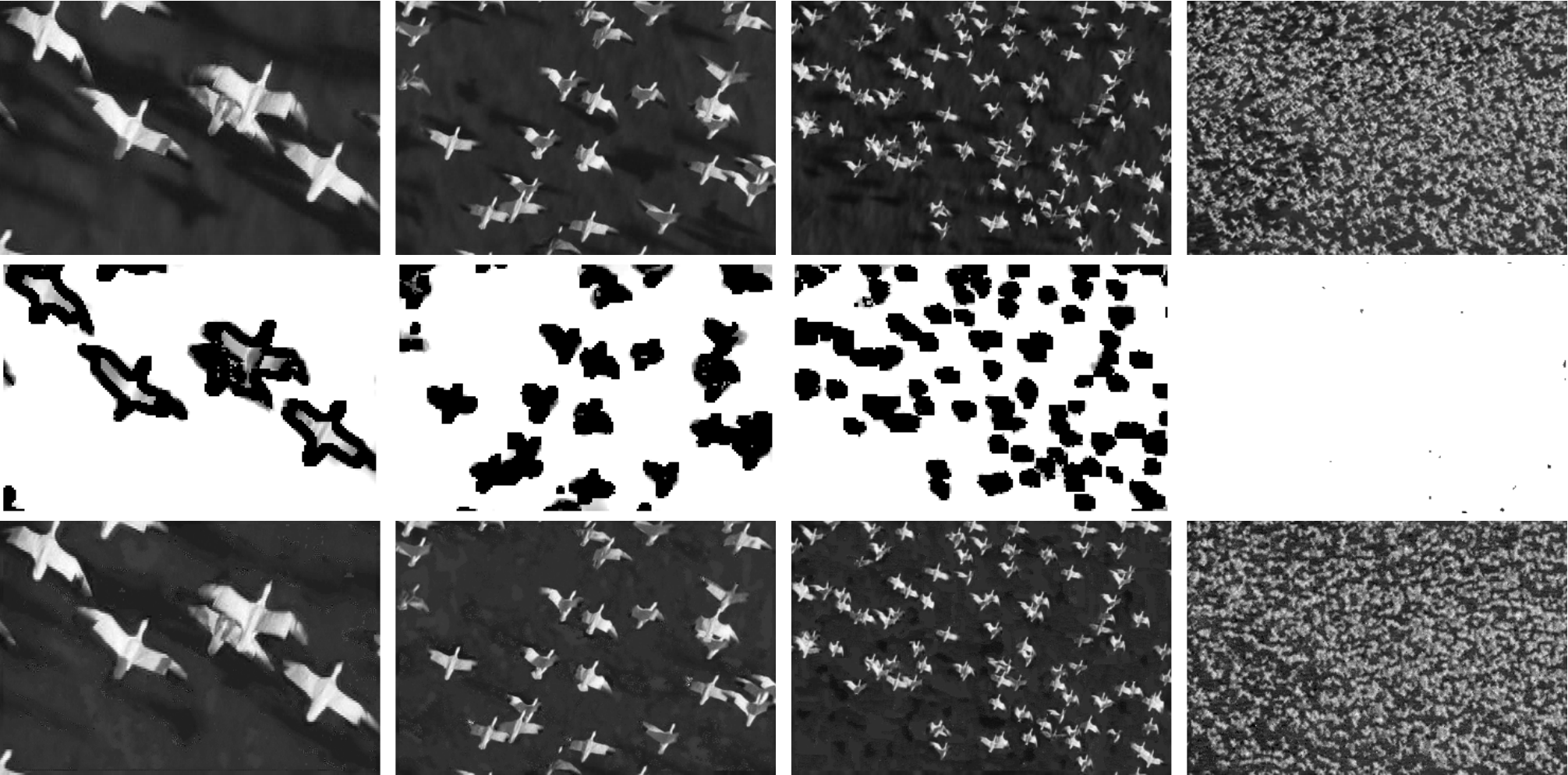}
\end{center}
   \caption{Representation switches triggered by scale. Row 1: observed frames; Row 2: trackability maps; Row 3: synthesized frames.}
\label{scale}
\end{figure}

\begin{figure}
\begin{center}
   \includegraphics[width=1\linewidth]{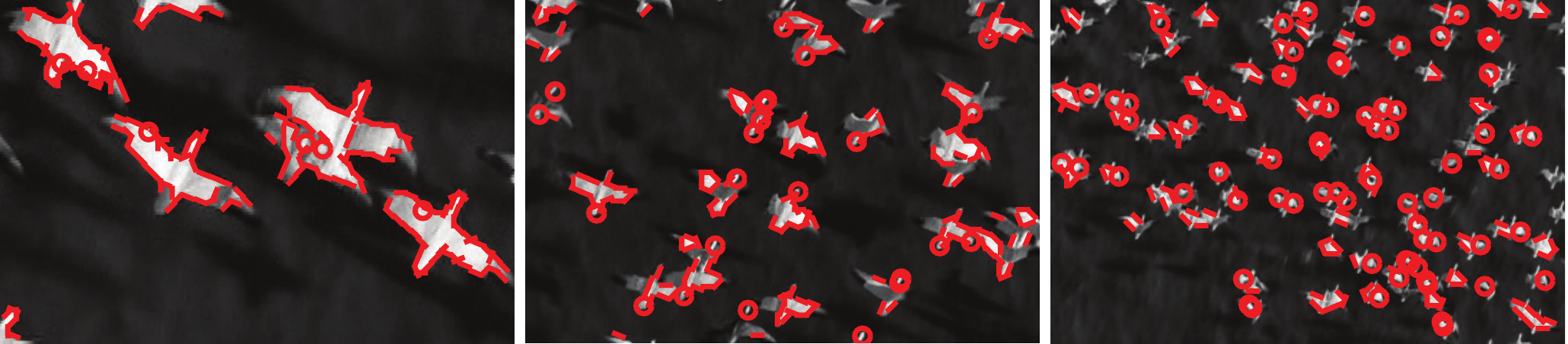}
\end{center}
   \caption{Representation types in different scale video frames, where circles represent blob-like type and short lines represent edge-like type.}\label{scale_rep}
\end{figure}

As it was observed in  (\cite{Gong2010}) that the optimal visual representation at a region is affected by distance, density and dynamics. In Fig.\ref{scale}, we show four video clips from a long video sequence. As the scale changes from high to low over time, the birds in the videos are perceived by lines of boundary, groups of kernels, dense points and dynamic textures respectively. We show the VPS of each clip and demonstrate that the proper representations are chosen by the model. Fig.\ref{scale_rep} shows the types of chosen primitives for explicit representations, in which circles represent blob-like type while short lines represent edge-like type primitives. Table \ref{rep_num} gives corresponding comparisons between the number of blob-like and edge-like primitives in each scale. For each scale, the comparison is within first 50, 100, 150 and 200 chosen primitives respectively. It is quite obvious that the percentage of chosen edge-like primitives in large scale frame is much higher than that in small scale. Meanwhile, in large scale frame, the blob-like primitives start to appear very late, which shows the fact that edge-like primitives are much more important in this scale for representing videos. But in small scale frame, the blob-like primitives possess a large percentage at the very beginning, and the number increase of edge-like primitives gets quicker and quicker while more and more primitives are chosen. This phenomenon demonstrates blob-like structures are much more prominent in small scale. So from this experiment, it is evident that VPS can choose proper representations automatically and furthermore, the representation patterns may reflect the scale of the videos.

\begin{table}[h]
\caption{The comparisons between the number of blob-like and edge-like primitives in 3 scales. For each scale, the numbers are compared in first 50, 100, 150 and 200 primitives respectively.}\label{rep_num}
\begin{center}
\begin{tabular}{|c|c|c|c|c|}
\hline
Scale & First 50 & First 100 & First 150 & First 200\\
\hline
1 & 0/50 & 0/100 & 1/149 & 6/194\\
\hline
2 & 0/50 & 6/94 & 16/134 & 23/177\\
\hline
3 & 19/31 & 37/63 & 58/92 & 71/129\\
\hline
\end{tabular}
\end{center}
\end{table}

\subsection{VPS supporting action representation}\label{highlevel}

\begin{figure}
\begin{center}
   \includegraphics[width=1\linewidth]{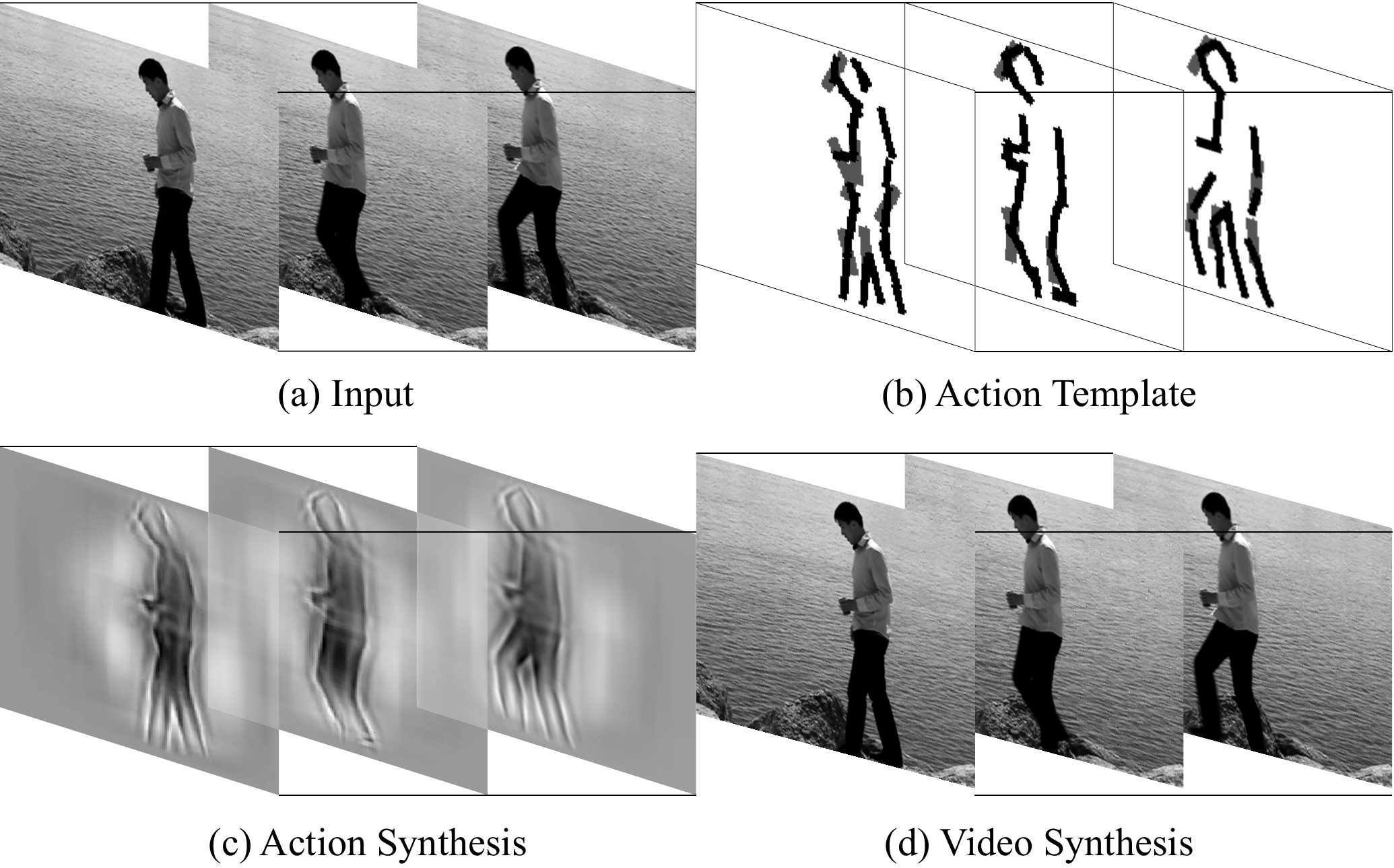}
\end{center}
   \caption{Action representation by VPS. (a) The input video. (b)
   Action template obtained by the deformable action
   template (\cite{Yao2009}). (c) Action synthesis by filters. (d)
   Video synthesis by VPS.}
\label{action}
\end{figure}

VPS is also compatible with high-level action representation. By grouping meaningful explicit parts in a principled way, it represents an action template. In Fig.\ref{action}, (b) is the action template given by the deformable action template model (\cite{Yao2009}) from the video shown in (a). The action template is essentially the sketches from the explicit regions. (c) shows an action synthesis with only filters from a matching pursuit process. While in (d), following the VPS model, the action parts and a few sketchable background are reconstructed by the explicit representation, and the large region of water is synthesized by the implicit representation; thus we get the synthesis of the whole video. Here, the explicit regions correspond to meaningful ``template'' parts, while the implicit regions are auxiliary background parts.

\begin{figure}
\begin{center}
   \includegraphics[width=1\linewidth]{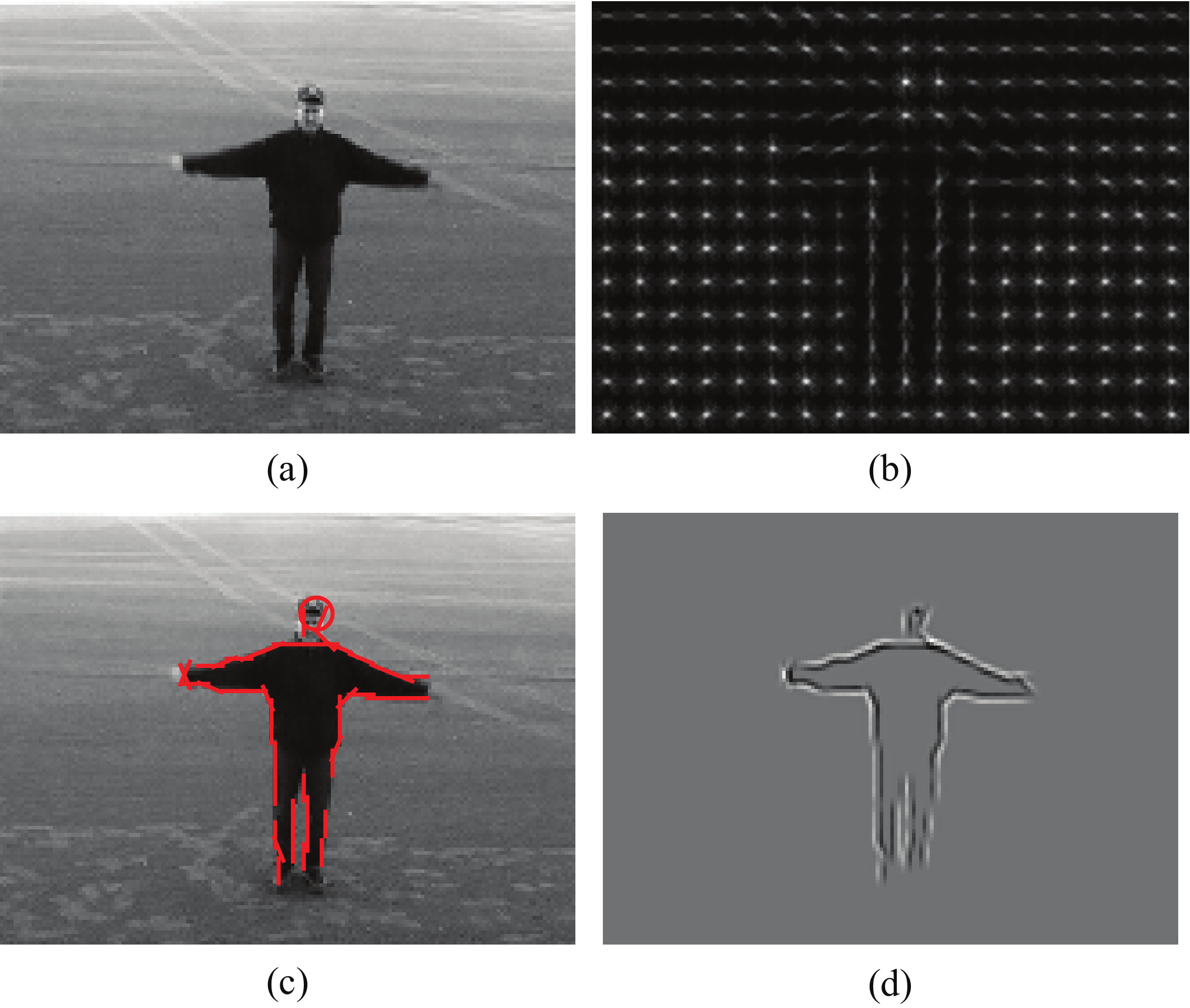}
\end{center}
   \caption{Structural information extracted by HOG and VPS. (a) The input video frame. (b)
   HOG descriptor. (c) VPS feature. (d) Boundary synthesis by filters.}
\label{vps_vs_hog}
\end{figure}

\begin{figure}
\begin{center}
   \includegraphics[width=1\linewidth]{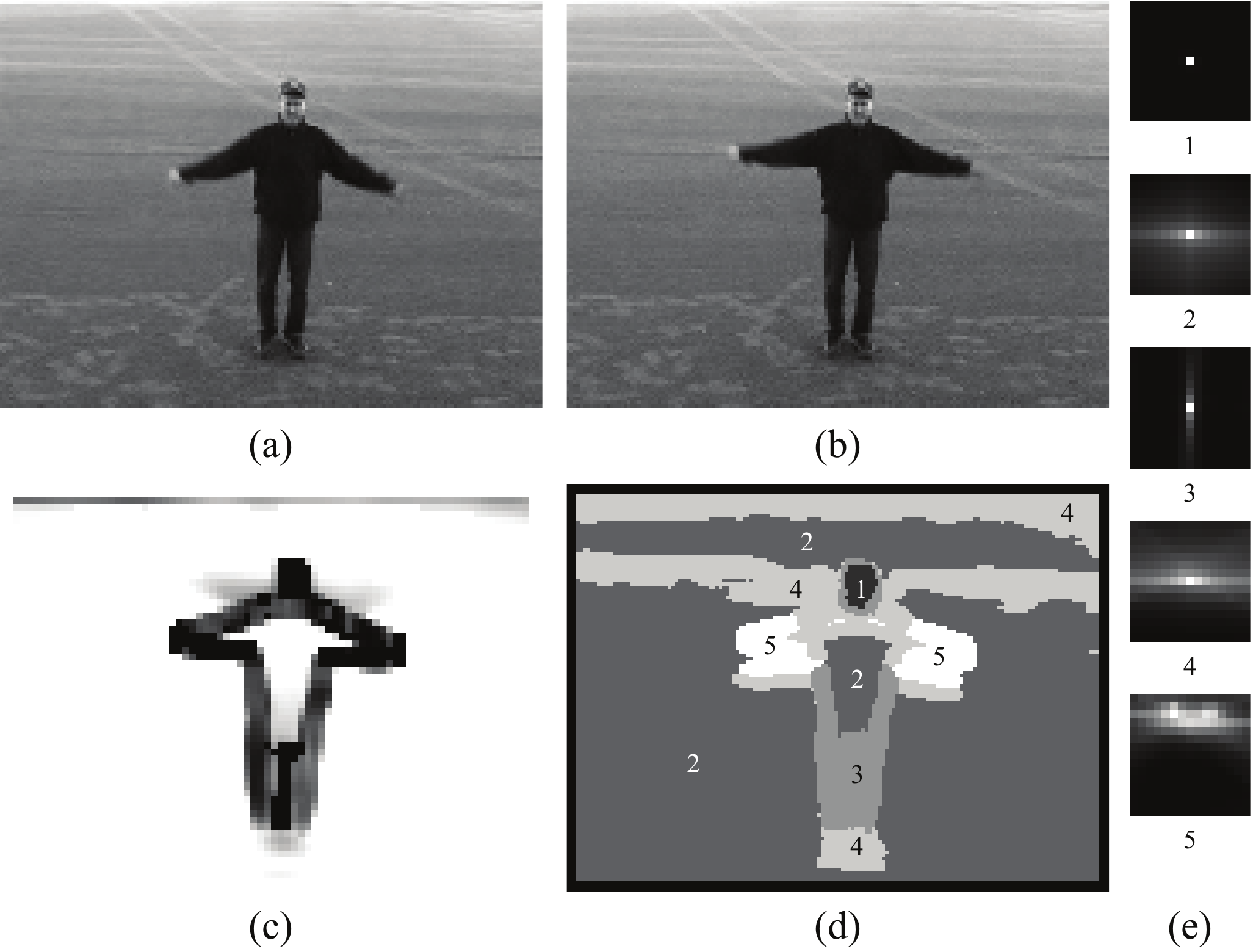}
\end{center}
   \caption{Motion statistics by VPS. (a) and (b)
   two continuous video frames of waving hands. (c) Trackability map. (d)
   Clustered motion style regions. (e) Corresponding motion statistics of each region.}
\label{action_seg}
\end{figure}

In order to show the relationship between VPS representation and effective high-level features, we take an KTH video (\cite{Schuldt2002}) as an example. Fig.\ref{vps_vs_hog} and Fig.\ref{action_seg} show the spatial and temporal features of explicit regions respectively. In Fig.\ref{vps_vs_hog}, we compare VPS spatial descriptor with well-known HOG feature (\cite{Dalal2005}), which has been widely used for object representation recently. (b) is the HOG descriptor for the human in one video frame (a). (c) shows structural features extracted by VPS, where circles and short edges represent 53 local descriptors. Compared with HOG in (b), VPS makes a local decision on each area based on statistics of filter responses, therefore it provides shorter coding length than HOG. Furthermore, it gives more precise description than HOG, e.g. the head part is represented by a circle descriptor, which contains more information than pure filter response histogram like HOG. And (d) gives a synthesis with corresponding filters, which shows the human boundary precisely.

In Fig.\ref{action_seg}, we show the motion information between two continuous frames (a) and (b) extracted by MA-FRAME in VPS. (d) gives the clustered motion styles in the current video. The motion statistics of the five styles are shown in (e) respectively. It is obvious that region 1 represents the area of head, which is almost still in the waving motion, while region 5 is for two arms, which shows definite moving direction. Region 3 represents the legs, which is actually an oriented trackable area. Region 2 and 4 are relatively ambiguous in motion direction, which are basically background of textures in the video. After giving the trackability map shown in (c) based on these motion styles, the motion template pops up.

In summary, the information extracted by VPS is compatible with high-level object and motion representations. Especially, it is very close to HOG and HOOF descriptors, which are proven effective spatial and temporal features respectively. The main difference is VPS makes a local decision to give a more compact expression and be better for visualization. Therefore, VPS does not only give a middle-level representation for video, but also has strong connection with low-level vision features and high-level vision templates.

\section{Discussion and Conclusion}

In this paper, we present a novel video primal sketch model as a middle-level generic representation of video. It is generative and parsimonious, integrating a sparse coding model for explicitly representing sketchable and trackable regions and extending the  FRAME models for implicitly representing textured motions. It is a video extension of the primal sketch model (\cite{Guo2007}). It can choose appropriate models automatically for video representation.

Based on the model, we design an effective algorithms for video synthesis, in which, explicit regions are reconstructed by learned video primitives and implicit regions are synthesized through a Gibbs sampling procedure based on spatio-temporal statistics. Our experiments shows that VPS is capable for video modeling and representation, which has high compression ratio and synthesis quality. Furthermore, it learns explicit and implicit expressions for meaningful low-level vision features and is compatible with high-level structural and motion representations, therefore provides a unified video representation for all low, middle and high level vision tasks.

In ongoing work, we will strengthen our work from several aspects, especially enhance the connections with low-level and high-level vision tasks. For low-level study, we are learning a much richer dictionary of $\Delta_B$ for video primitives, which is more comprehensive. For high-level application, we are applying the VPS features to object and action representation and recognition.

\begin{acknowledgements}
This work is done when Han is a visiting student at UCLA. We thank the support of an NSF grant DMS 1007889 and ONR MURI grant N00014-10-1-0933 at UCLA. The authors also thank the support by four grants in China: NSFC 61303168, 2007CB311002, NSFC 60832004, NSFC 61273020.
\end{acknowledgements}

\bibliographystyle{spbasic}      
\bibliography{CV_bib}   

%
%

\end{document}